\documentclass{article} 
\usepackage[final]{colm2026_conference}

\usepackage{microtype}
\usepackage{hyperref}
\usepackage{url}
\usepackage{booktabs}
\usepackage{amsmath}
\usepackage{mathtools}
\usepackage{nicematrix}
\usepackage{caption}
\usepackage{subcaption}
\usepackage{algorithm}
\usepackage{algpseudocode}
\usepackage{tcolorbox} 

\usepackage{cleveref}
\usepackage{xspace}
\usepackage{lineno}

\usepackage{pgfplots}
\pgfplotsset{compat=1.18}
\usepackage[edges]{forest}
\definecolor{paired-light-blue}{RGB}{198, 219, 239}
\definecolor{paired-dark-blue}{RGB}{49, 130, 188}
\definecolor{paired-light-orange}{RGB}{251, 208, 162}
\definecolor{paired-dark-orange}{RGB}{230, 85, 12}
\definecolor{paired-light-green}{RGB}{199, 233, 193}
\definecolor{paired-dark-green}{RGB}{49, 163, 83}
\definecolor{paired-light-purple}{RGB}{218, 218, 235}
\definecolor{paired-dark-purple}{RGB}{117, 107, 176}
\definecolor{paired-light-gray}{RGB}{217, 217, 217}
\definecolor{paired-dark-gray}{RGB}{99, 99, 99}
\definecolor{paired-light-pink}{RGB}{222, 158, 214}
\definecolor{paired-dark-pink}{RGB}{123, 65, 115}
\definecolor{paired-light-red}{RGB}{231, 150, 156}
\definecolor{paired-dark-red}{RGB}{131, 60, 56}
\definecolor{paired-light-yellow}{RGB}{231, 204, 149}
\definecolor{paired-dark-yellow}{RGB}{141, 109, 49}
\definecolor{light-green}{RGB}{118, 207, 180}
\definecolor{raspberry}{RGB}{228, 24, 99}

\definecolor{darkblue}{rgb}{0, 0, 0.5}
\hypersetup{colorlinks=true, citecolor=darkblue, linkcolor=darkblue, urlcolor=darkblue}

\newcommand{\method}{Fork-think\xspace} 

\definecolor{jgreen}{RGB}{60, 163, 83}
\definecolor{zblue}{RGB}{49, 130, 188}

\DeclareMathOperator*{\argmax}{arg\,max}
\DeclareMathOperator*{\argmin}{arg\,min}

\algnewcommand\algorithmicforeach{\textbf{for each}}
\algdef{S}[FOR]{ForEach}[1]{\algorithmicforeach\ #1\ \algorithmicdo}

\title{Fork-Think with Confidence}

\author{Zena Al-Khalili, Rafi Hakim, Dietrich Klakow$^*$, Ji-Ung Lee\thanks{Co senior authors.} \\ 
Saarland Informatics Campus, Saarland University, Germany \\
\texttt{zakhalili@lsv.uni-saarland.de} 
}

\begin{document}

\ifcolmsubmission
\linenumbers
\fi

\maketitle

\begin{abstract}
Parallel thinking has enjoyed great success for boosting LLM performance on reasoning tasks without the need for any re-training.
However, existing methods follow a \textit{think-first-then-decide} paradigm, i.e., they first sample multiple reasoning paths, which inevitably leads to overgeneration, then prune or stop unnecessary paths to compensate.
In contrast, \textit{decide-first-then-think}---i.e., first identifying points that are likely to lead to desirable generations---has been underexplored so far.
Following this paradigm, we propose \method with confidence, that first identifies forking points using model confidence in a single seeding path, then triggers thinking, sampling multiple continuations and aggregating them for the final response. 
Our experiments across three models and three reasoning benchmarks show that \method reduces the token consumption by up to 30\% and run-time by up to 57\%, while performing comparable or better to parallel thinking. 
Our analysis reveals that \method is able to identify forking points that are meaningful with respect to the downstream task and that sampling at later positions can lead to substantially better generations.
Finally, we demonstrate how combining \method with existing mechanisms such as early stopping and weighted voting can further boost the performance and perform comparably to existing state-of-the-art methods, without requiring any warm-up or offline training.
Our results establish pre-determined forking as a promising research direction for efficient LLM reasoning.\footnote{Code available at:  \url{https://github.com/uds-lsv/Fork-Think-with-Confidence}}

\end{abstract}

\section{Introduction}\label{sec:01-introduction}

Despite the recent success of large language models (LLMs), they still struggle to reliably solve complex tasks, especially those that require multi-step reasoning and deduction~\citep{wang2024mmlupro}. 
To combat this, research has investigated various methods for building advanced reasoning models, for instance, by running 
supervised fine-tuning on additional reasoning data~\citep{chung2024scaling, wang2024openropensourceframework} or using reinforcement learning~\citep{guo2025deepseek}. 
However, such post-training methods are limited by the available data and can require substantial amounts of compute~\citep{rafailov2023dpo, kumar2025llmposttrainingdeepdive}.

In contrast, \emph{inference-time scaling} avoids costly post-training (or can be used in a complementary manner) by allocating additional compute during inference~\citep{zhao2025scalinglaw, welleck2024from}.
Over time, works have devised sequential methods that iteratively refine the quality of the generation~\citep{ni2023learning} as well as search-based methods that sample (and aggregate) multiple continuations at each step in the generation~\citep{yao2023tree}.
However, sequential methods are bound by the maximum context size or deteriorate with overlong contexts~\citep{liu-etal-2024-lost} and search-based methods incur an exponential compute cost. 
To reduce compute cost, works have proposed \emph{parallel thinking} as a lightweight alternative, where multiple reasoning paths were sampled only once---allowing compute cost to scale linearly with the number of samples---and then aggregated~\citep{wang2023selfconsistency, brown2024largelanguagemonkeysscaling}.
Since then, researchers have studied increasingly sophisticated aggregation methods~\citep{wang-etal-2025-ranked, kang2025scalable} or even proposed to break up the reasoning process again into smaller steps~\citep{inoue2025wider}, however, at the expense of an increased compute cost especially for verbose reasoning models~\citep{balachandran2025inference}.
Most recently, some works proposed to mitigate this using adaptive methods that dynamically identify and prune unpromising reasoning traces using a threshold found in a warm-up phase~\citep{fu2026deep} or a scoring model trained offline~\citep{liang2026hiddenstatesearlysignals,tu2025deeppruneparallelscalingintertrace}.

One common denominator of prior works is that they operate on a \emph{think-first-then-decide} paradigm (i.e., sampling at the beginning), without a theoretical justification to do so. 
In this work, we propose to address parallel thinking in a \emph{decide-first-then-think} manner which has not been studied so far. 
In other words, we propose to first identify \emph{fork}ing points using a seed path from which we then sample \emph{think}ing paths (\method). 
To identify forking points, we utilize pivot tokens which have been found to substantially affect the generation~\citep{abdin2024phi4technicalreport} and which we identify using the model's confidence.
Our contributions are:
\begin{enumerate}
    \item A new perspective on parallel thinking based on a \emph{decide-first-then-think} paradigm and a first method (\method) based on the novel paradigm.
    \item A systematic evaluation across three models and three datasets showing that \method is a simple, yet effective method reducing the overall token cost by up to 30\% and the overall run-time by up-to $\sim$57\% in comparison to parallel think.  
    \item Analysis and ablation studies investigating different factors and showing that later forking points can achieve a substantially higher performance compared to generating parallel samples at the beginning.
\end{enumerate}

Our results indicate that the \emph{decide-first-then-think} paradigm provides a promising new perspective on parallel scaling and that investigating more sophisticated methods to identify forking points could be a worthwhile endeavor for future research. 

\begin{figure}[!tb]
     \centering
     \begin{subfigure}[b]{0.40\textwidth}
         \centering
         \includegraphics[width=\textwidth]{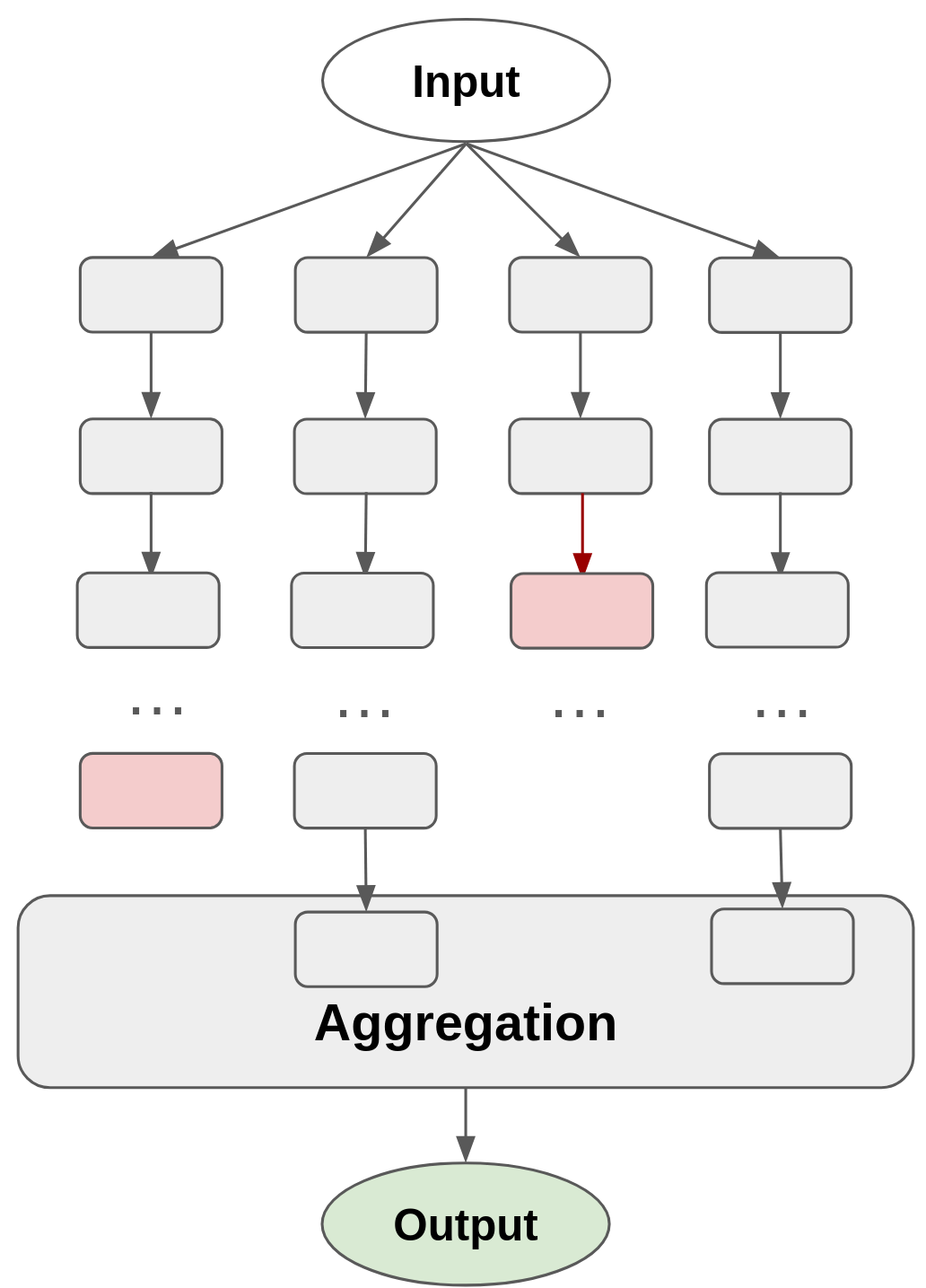}
         \caption{\textit{Think-first-then-Decide}}
         \label{fig:01-think-first-then-decide}
     \end{subfigure}
     \hfill
     \begin{subfigure}[b]{0.40\textwidth}
         \centering
         \includegraphics[width=\textwidth]{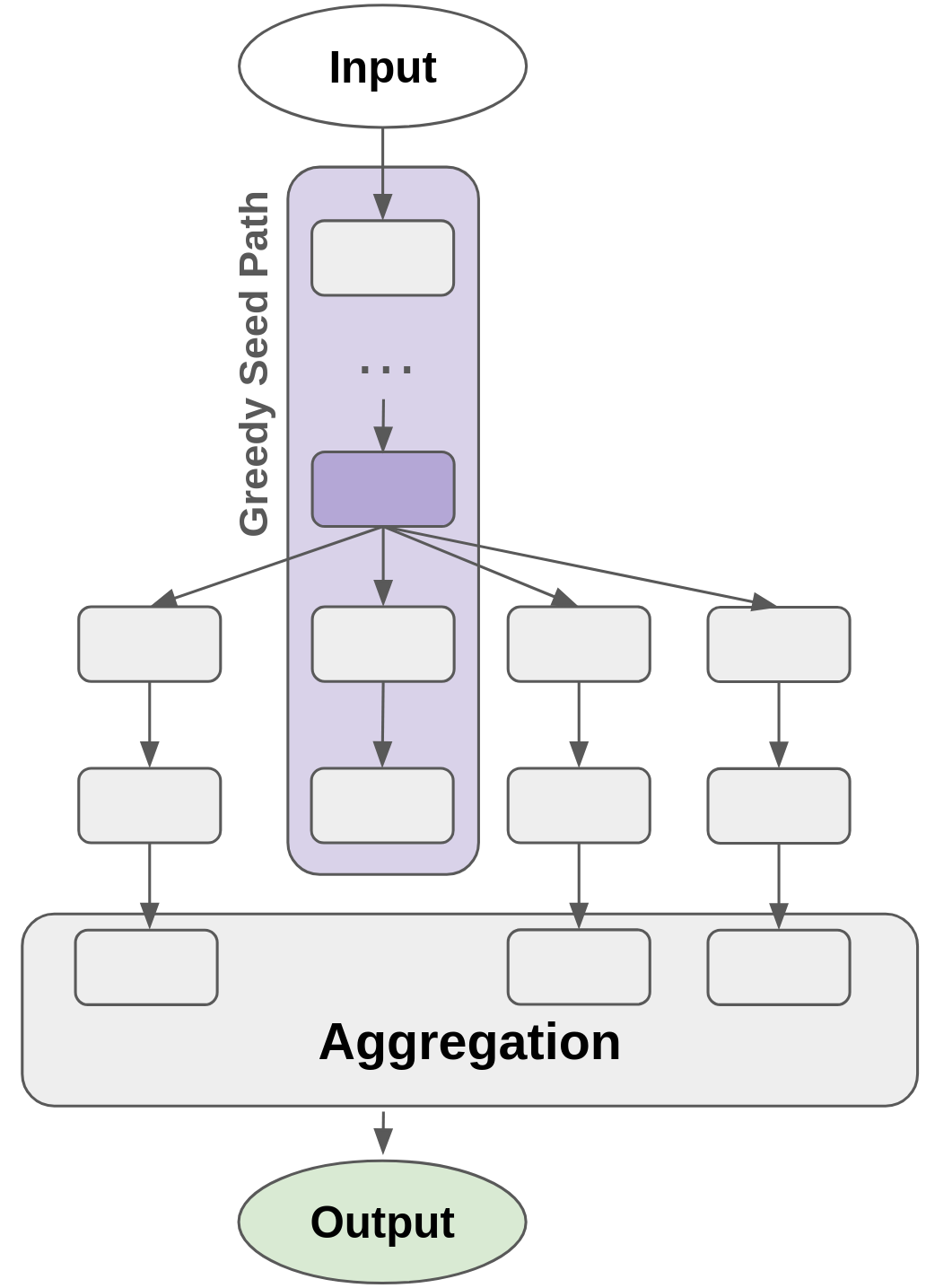}
         \caption{\textit{Decide-first-then-Think}}
         \label{fig:01-decide-first-then-think}
     \end{subfigure}
        \caption{\textbf{Left}: Existing works  commonly pursue a \emph{think-first-then-decide} strategy  which given a specific input (i.e., prompt \& question), first generates multiple parallel traces which are pruned dynamically and aggregated afterwards. \textbf{Right}: Instead, we propose a \emph{decide-first-then-think} strategy, which generates a greedy seed path first, and then identifies promising forking points to sample multiple parallel traces from. Note, that \method can be used complementary to existing pruning or aggregation methods.}
        \label{fig:01-paradigm}
\end{figure}

\section{Related Work}\label{sec:02-rel-work}

\subsection{LLM Reasoning}\label{rel-work-adaptive}

Evaluating and improving the reasoning capabilities of LLMs has been a long-sought subject of research~\citep{takeshi2022zeroshot,wang-etal-2023-chatgpt-defend, suzgun-etal-2023-challenging, al2025evaluating, liu2025pricinglogic}. 
Whereas early works were primarily concerned with improving the reasoning capabilities via in-context learning (ICL, \citealt{xie2022an}) or chain-of-thought prompting (CoT, \citealt{wei2022cot}), recent works have also investigated the verification of reasoning chains~\citep{lightman2024lets,hosseini2024vstar,feng2026vericot} as well as how to directly instill reasoning capabilities into models~\citep{guo2025deepseek, setlur2025rewarding}. 
Others propose to break down a complex task into easier sub-tasks~\citep{zhou2023leasttomost} or to minimize randomness (e.g., by falling back to greedy decoding) at critical reasoning steps to avoid wrong deductions~\citep{troshin2025control}. 

\paragraph{The Importance of Diversity.} 
One recurring property to improve LLM reasoning is the diversity of the generated reasoning paths (for a single question).
For instance, \citet{li-etal-2023-making} show that aggregating responses with diverse reasoning paths (using different prompts) for a single question substantially boosts performance.  
Others propose to increase diversity by ensembling~\citep{dang2025weight} or by sampling from a different distribution~\citep{karan2025reasoningsamplingbasemodel}.
Finally, some methods directly mitigate the diversity collapse caused by post-training~\citep{song2025outcomebased}, by adaptively changing the temperature~\citep{wu2025roletemperaturesamplingtesttime, yue2025doesreinforcementlearningreally} or adding regularizers during post-training~\citep{he-etal-2025-rewarding, ju2026reasoningpathdivergencenew}.
However these methods often rely upon repeatedly sampling the model at different temperatures (or require more offline training) which substantially increases compute cost.
Instead, we propose to utilize pivot tokens~\citep{abdin2024phi4technicalreport} as our forking points. 

\subsection{Inference-time Scaling}

Inference-time scaling (or test-time scaling) allows models to dynamically adjust their generation distribution during inference and has emerged as a powerful alternative to scaling model parameters~\citep{snell2025scaling, wu2025inference}. 
While scaling strategies such as sequential refinement~\citep{madaan2023selfrefine} or search against a verifier~\citep{wu2025inference} are effective, they are limited to single-path exploration or rely on expensive, difficult-to-train scoring functions.
In contrast, parallel thinking (or repetitive sampling) bypasses the need for trained verifiers by independently sampling multiple, diverse reasoning paths via temperature scaling~\citep{wang2023selfconsistency, brown2024largelanguagemonkeysscaling}.
These are then aggregated using voting strategies such as majority voting~\citep{wang2023selfconsistency} and confidence-weighted voting~\citep{taubenfeld-etal-2025-confidence, kang2025scalable}, or re-ranking techniques~\citep{leang2026picsar}.
While parallel thinking yields substantial performance gains on complex reasoning tasks, it incurs a prohibitive token cost, especially for the long CoT generations of reasoning models~\citep{guo2025deepseek}.

\paragraph{Adaptive Methods.}
To alleviate the computational overhead of parallel thinking using static generation or aggregation schemes, works have investigated methods that allocate compute more adaptively in a training-free manner~\citep{wu2026efficiencyadaptivitydeeperlook}. 
Prior approaches typically follow pruning-based strategies that halt some traces based on a model's confidence~\citep{fu2026deep} or signals from hidden states~\citep{liang2026hiddenstatesearlysignals}.
Others halt the sampling process when a consensus is reached within a predefined window of samples~\citep{li2024escape} or the majority answer generated so far is statistically undeniable~\citep{aggarwal-etal-2023-lets}.
Some even investigate using offline training with additional objectives such as reducing redundancy~\citep{tu2025deeppruneparallelscalingintertrace}. 
Albeit increasing the efficiency, we argue that such a think-first-then-decide approach might lead to suboptimal branching points, and instead propose a decide-first-then-think view of parallel thinking.
Finally, a very recent work which mixes both paradigms is EAGER~\citep{scalena2026eager}. 
Here, the authors propose to adaptively branch during generation based on a threshold that requires a warm-up phase with an extensive hyper-parameter search. 
In contrast, \method does not require any warm-up phase or offline training.

\section{Fork-think with Confidence}\label{sec:03-method}

Existing methods primarily utilize a \emph{think-first-then-decide} paradigm, i.e., they first sample multiple reasoning paths and then decide which to consider for the final response.
This inherently leads to an overgeneration of reasoning paths, making pruning or early stopping essential and resulting in offline training or warm-up.
We instead propose to shift the paradigm towards \emph{decide-first-then-think}: we first generate a single seed path and then identify \emph{forking points} from which we further sample multiple reasoning paths with a temperature $\tau > 0$. 
Note that we do not require the seed path to be complete (i.e., a partial response suffices) which is distinctly different from sequential inference-time scaling methods that iteratively improve upon a single generated response.

\subsection{Identifying Forking Points}\label{sec03:forking-tokens}

Ideally, we would sample from forking points that maximize the chances of producing correct responses. 
However, this is not possible due to the absence of a scoring function which would require an offline training or warm-up phase.
Instead, we  use previous findings in LLM reasoning that show how aggregating a diverse set of responses generated for the same question can improve overall performance~\citep{li-etal-2023-making}.
Hence, our objective is to identify the points where sampling is likely to generate diverse responses.

\paragraph{Pivot Tokens.}
Multiple works report the presence of specific tokens or token windows that have a substantial impact on the future generation (positively or negatively).
These so-called pivot tokens (sometimes also referred to as forking or critical tokens) typically correspond to a lower \emph{model confidence} at a specific position in the generation process. 
Whereas previous work primarily utilizes these pivot tokens to improve LLM post-training~\cite{abdin2024phi4technicalreport, wang2025beyond}, we propose to utilize them as forking points where sampling with a temperature $\tau > 0$ is likely to generate diverse responses.
To quantify the confidence of the model $c_i$ at the $i$-th position, we follow prior work and utilize the log-probability averaged over the $k$ most likely tokens $x_{i1},\ldots, x_{ik}$:
\begin{equation}
   c_i = \frac{1}{k}\sum_{j=1}^{k}\log P(x_{ij}), \label{eq:confidence}
\end{equation}
Where $P(x_{ij})$ are the individual token probabilities of the original model normalized over $k$. 
The best forking point $c^*$ for a string $X$ and the respective model confidences $C$ is then described as:
\begin{equation}
   c^* = \argmin_{x_i \in X} C,  \label{eq:least-confidence}
\end{equation}
i.e., the point where the model confidence is lowest.\footnote{While we use model confidence, we note that there exist other metrics for quantifying a model's uncertainty~\citep{tao2025revisiting}. We provide respective ablation studies in \cref{sec:06-results}.}
Note that \cref{eq:least-confidence} can easily be extended to multiple forking points by sorting all tokens $x_i \in X$ according to their respective model confidences $c_i$ and selecting the ones where the model is least confident.

\subsection{\method}
Having defined the selection criteria to identify forking points, we now define \method. 
We first specify the length $l_0$ of the seed path, the number of top-$k$ tokens to estimate the confidence of the model, a branching factor $n$ that determines how many samples are generated at the forking point, the number of forking points $m$ to sample from, and the temperature $\tau$ to use for sampling (after seed path generation).
Following \citet{abdin2024phi4technicalreport}, we utilize greedy sampling (i.e., $\tau_0 = 0$) to generate a seed path $X = \{x_1, \ldots, x_l\}$ with length $l_0$ and determine top-$m$ forking points $C^* = \{c_1, \ldots, c_m \}$ according to \cref{eq:least-confidence}. 
Next, we generate $n$ samples for each forking point $c_j \in C^*$ using a prefix $\pi_j = \{x_1, \ldots, x_{<c_j}\}$. 
Finally, we aggregate all generated samples through majority voting.
\Cref{alg:fork-think} provides an overview of the complete pipeline of \method. 

\begin{algorithm}[t]
\begin{small}
\caption{Fork-think with Confidence}
\label{alg:fork-think}
\begin{algorithmic}[1]

\Require prompt $p$, question $q$
\Ensure seed path length $l_0$, top-$k$, branching factor $n$, number of forking points $m$, temperature $\tau$ 

\State $X \gets \textsc{Generate}(p, q; \tau_0)$ \Comment{seed path}
\State $C \gets \emptyset$
\For{$i = 1$ to $l_0$}
    \State $c_i = \frac{1}{k}\sum_{j}^{k}\log P(x_{ij})$
    \State $C \gets C \cup \{c_i\} $
\EndFor
\State $\Pi \gets \emptyset$
\For{$j = 1$ to $m$}
\State $c^* = \argmin_{x_i \in X} C$ \Comment{forking point}
\State $\pi_j = \{x_{1}, \ldots, x_{<c^*}\}$ \Comment{generate prefix}
\State $\Pi \gets \Pi \cup \{\pi_j\}$
\State $C \gets C \setminus \{c^*\} $
\EndFor
\State $\mathcal{A} \gets \emptyset$ \Comment{define answer set}
\ForEach{$\pi \in \Pi$}
\For{$i = 1$ to $n$}
    \State $y_i \gets \textsc{Generate}(p, q, \pi; \tau)$ \Comment{sample response}
    \State $a_i \gets \textsc{ExtractAnswer}(y_i)$
    \State $\mathcal{A} \gets \mathcal{A} \cup \{a_i\}$
\EndFor
\EndFor
\State $\mathcal{V} \gets \textsc{CountVotes}(\mathcal{A})$ \Comment{majority-voting}
\State $a^* \gets \argmax_{a} \mathcal{V}[a]$
\State \Return $a^*$
\end{algorithmic}
\end{small}
\end{algorithm}

\section{Evaluation}\label{sec:05-experiments}

Our experiments comprise three reasoning benchmarks, comparing \method against five baselines and state-of-the-art methods across three different models.
Our evaluations also include Flex-\method, a variation of \method augmented with early stopping and weighted majority voting.
As the run-time analysis in appendix \ref{sec:appendix-runtime} shows, \method leads to substantially shorter run-times compared to Parallel Thinking.

\subsection{Experimental Setup}
We quantify the performance of each method using majority-voting accuracy (in \%), the number of tokens used (in million).

\paragraph{Models.}
We evaluate \method using three models from different model families, Qwen3-8B~\citep{yang2025qwen3technicalreport}, DeepSeek-R1 distilled into a Qwen3-8B base model (referred to as DeepSeek-8B, \citealt{guo2025deepseek}), and Phi-4 Reasoning Plus 14B (Phi-4-RP-14B, \citealt{abdin2025phi4reasoningtechnicalreport}). 
For each model, we set the hyper-parameters (maximum trace length, temperature $\tau$, top-$p$) to the ones recommended in the respective model card. 
We provide all details of the model-specific hyper-parameters in appendix~\ref{sec:appendix-parameters} and all prompt templates in appendix~\ref{sec:appendix-prompts}.

\paragraph{Benchmarks.}
Following previous works, we use two widely-used benchmarks for evaluating LLM reasoning which are high-difficulty mathematical competition problems also featured in the MathArena leaderboard~\citep{balunovic_srimatharena_2025}.
The problems stem from AIME (American Invitational Mathematics Examination), which are exams used to select the U.S. representative students at the international math olympiad. 
Each year consist of two rounds of exams (I \& II) with each exam consisting of 15 challenging questions. We use the questions from the years 2024 (AIME24, \citealt{AIME24I,AIME24II}) and 2025 (AIME25, \citealt{AIME25I,AIME25II}). 
To benchmark \method beyond mathematical reasoning, we also run experiments on GPQA~\citep{rein2024gpqa}, a dataset consisting of graduate-level scientific questions. 
More specifically, we use the diamond split, which consists of 198 questions answered correctly by experts and incorrectly by non-experts.

\paragraph{Baselines.}
We evaluate \method against several baselines as well as a very recent state-of-the-art method to assess its practical effectiveness.

\begin{description} 
    \item[Greedy] Our first baseline reports the performance of greedy decoding ($\tau = 0$) without any parallel sampling. Instead, we continue generating from the greedy seed path with an extended length that matches the maximum trace length granted to the other methods. This serves as a lower bound on \method's performance.
    \item[CoT] Our second baseline is a single reasoning trace that follows standard Chain-of-Thought prompting \citep{wei2022cot}.
    \item[Parallel Thinking] Our primary baseline was proposed by \citet{wang2023selfconsistency} and is a widely used baseline in LLM reasoning tasks. Starting from the given prompt (i.e., the instruction and a question), it first generates $n$ parallel reasoning traces, then aggregates the responses using a self-consistency mechanism (i.e., majority voting) to obtain the final response. In our experiments, we refer to this method as parallel thinking.
    \item[ASC] We also compare to Adaptive Self-Consistency (ASC, \citealt{aggarwal-etal-2023-lets}) that generates reasoning traces one by one, and models the probability distribution over generated answers to quantify the confidence in the majority one, then stops the generation when the majority answer becomes statistically undeniable.
    \item[DeepConf] Finally, we include a very recently published state-of-the-art method, namely, Deep Think with Confidence (DeepConf, \citealt{fu2026deep}). DeepConf generates parallel reasoning traces which are pruned dynamically (i.e., during generation) using a threshold that is estimated during a warm-up phase. The generated traces are then aggregated using a confidence-weighted majority voting scheme. For a fair comparison, we report the performance for both aggregation schemata (confidence-weighted and non-weighted majority voting).\footnote{Note, that between EAGER~\citep{scalena2026eager} and DeepConf, DeepConf is the stronger baseline as reported by \citet{scalena2026eager}.}
   
\end{description}

\paragraph{Other Hyper-parameters.}
For evaluation, we use a seed path length $l_0=2,048$, a branching factor $n=32$ (also for Parallel Thinking and DeepConf), and set the number of forking points $m=1$.
For the main experiments, we specifically use only a single forking point to better quantify the effect of branching later (\method) compared to branching from the very beginning (Parallel Thinking and DeepConf). 
We further set top-$k=10$ and select the branching temperature $\tau$ as per recommendation in the respective model card (cf. appendix~\ref{sec:appendix-parameters}).
Trace aggregation is done via simple majority voting for Parallel Thinking and \method.
For DeepConf, we perform a confidence-weighted majority voting using the best performing confidence metric reported by the authors.

\paragraph{Implementation Backend.}
We implement Parallel Thinking, Greedy, and \method using SGLang~\citep{zheng2024sglangefficientexecutionstructured} as our backend. Fork-think heavily depends on prefixing, which is more compatible with SGLang, given its radix cache that skips massive amounts of prefill compute if a request shares a context history with prior requests.
For DeepConf, we use the implementation provided by the authors that uses vLLM as the backend~\citep{kwon2023vllm}.

\subsection{Results}

\begin{table}[t]
  \centering
  \small
  \begin{NiceTabular}{lcccccccccc}
  \toprule
  \textbf{Model} 
  & \multicolumn{2}{c}{\textbf{Greedy}} 
  & \multicolumn{2}{c}{\textbf{CoT}} 
  & \multicolumn{2}{c}{\textbf{Parallel Thinking}} 
  & \multicolumn{2}{c}{\textbf{\method}}  \\
  \cmidrule(lr){2-3} \cmidrule(lr){4-5} \cmidrule(lr){6-7} \cmidrule(lr){8-9} 
   & \textbf{Token}  & \textbf{Acc} 
   & \textbf{Token}  & \textbf{Acc} 
   & \textbf{Token} & \textbf{Acc}  
   & \textbf{Token ($\Delta$\%)} & \textbf{Acc}  \\
  \midrule
  
  \rowcolor{paired-light-purple} \Block{1-9}{\textbf{AIME24}} \\
  \midrule
  Qwen3-8B     & 0.70 & 50.0 & 0.6 & 68.7 & 20.2 & 80.1  & 14.2 (-30\%) & \textbf{81.8}   \\
  DeepSeek-8B  & 0.68 & 80.0 & 0.7 & 80.0 & 22.6 & \textbf{86.0}  & 20.3 (-10\%) & \textbf{86.0}  \\
  Phi-4-RP-14B & 0.73 & 43.3 & 0.4 & 67.3 & 12.2 & \textbf{77.8} & 11.2 (-8\%)  & 77.0  \\
  \midrule
  
  \rowcolor{paired-light-purple} \Block{1-9}{\textbf{AIME25}} \\
  \midrule
  Qwen3-8B     & 0.62 & 60.0 & 0.7 & 64.0 & 22.0  & 75.4 & 15.9 (-27\%) & \textbf{75.6}  \\
  DeepSeek-8B  & 0.75 & 80.0 & 0.8 & 75.3 & 25.2 & \textbf{83.3}  & 21.4 (-14\%) & 82.4 \\
  Phi-4-RP-14B & 0.66 & 46.6 & 0.4 & 58.7 & 13.5  & 62.6  & 12.5 (-7\%) & \textbf{66.6}  \\
  \midrule
\rowcolor{paired-light-purple} \Block{1-9}{\textbf{GPQA-Diamond}} \\
  \midrule
Qwen3-8B     & 3.3 & 39.8 & 2.6 & 55.2 & 83.3 & 62.1 & 60.4 (-27\%) & \textbf{63.2}  \\
DeepSeek-8B  & 2.3 & 61.1 & 2.5 & 60.8 & 72.4 & \textbf{66.9} & 54.2 (-25\%) & 62.1  \\
Phi-4-RP-14B & 3.6 & 31.3 & 1.7 & 53.9 & 54.9 & 67.3 & 53.1(-3\%) & \textbf{70.1}  \\

  \bottomrule
  \end{NiceTabular}
  \caption{Accuracy (in \%) and required tokens (in $\times 10^{6}$) for a branching factor $n=32$. All results are aggregated using majority voting. For \method, we set the seed path length $l_0=2,048$ and also include it as additional cost in the required token computation.}
  \label{tab:main_res}
\end{table}

\Cref{tab:main_res} shows the performance of \method and baselines that do not utilize pruning or weighted aggregation methods beyond majority voting.
We repeat each experiment across all models five times using different seeds for the pseudo-random number generator.
We find that \method uses substantially fewer tokens than Parallel Thinking, while achieving a comparable overall accuracy.
Moreover, we can see that the improvements in terms of overall token usage as well as the performance are maintained across all models and datasets. We can also see that scaling the compute at inference time substantially outperforms the single-sample variants (greedy and CoT), highlighting the benefits of sampling multiple times instead of scaling model parameters.

\paragraph{Flex-\method.} 
One advantage of \method is that it is easily extensible with more sophisticated methods such as early stopping and other weight aggregation methods.
To provide a fair comparison against methods that utilize such mechanisms, we evaluate Flex-\method, a variant of \method that includes confidence-weighted majority voting and early stopping.
The early stopping strategy is similar to the one introduced in ASC, i.e., after forking, the samples are generated one by one until the majority answer is statistically prevalent.  
\Cref{tab:add_res} shows the performance of Flex-\method against DeepConf and ASC. 
To provide a fair comparison against ASC which does not rely on confidence-weighted majority voting (WMV), we also report the majority voting (MV) scores for DeepConf and Flex-\method.
Overall, we find that Flex-\method achieves a comparable (or slightly lower) performance compared to DeepConf; however, with substantially less tokens.
Moreover, we observe that it performs on-par with ASC also in terms of accuracy.
Interestingly, we find again that Flex-\method consistently outperforms all other methods for the weaker (Qwen3-8B) model.
Overall, our results suggest that selecting a later forking point can already save a substantial number of tokens while achieving a comparable performance, showing that \emph{decide-first-then-think} is a promising paradigm.

\begin{table}[t]
  \centering
  \small
  \begin{NiceTabular}{lccccccc}
  \toprule
  \textbf{Model} 
  & \multicolumn{2}{c}{\textbf{ASC}} 
  & \multicolumn{2}{c}{\textbf{DeepConf}} 
  & \multicolumn{2}{c}{\textbf{Flex-\method}} \\
  \cmidrule(lr){2-3} \cmidrule(lr){4-5} \cmidrule(lr){6-7}
   & \textbf{Token} & \textbf{MV} 
   & \textbf{Token } & \textbf{MV} (WMV)  
   & \textbf{Token } & \textbf{MV} (WMV) \\
  \midrule
  
\rowcolor{paired-light-purple} \Block{1-7}{\textbf{AIME24}} \\
\midrule
Qwen3-8B     & 10.4 & 80.1  & 11.1 & \textbf{81.8} (\(81.2\)) & \textbf{7.0} & \textbf{81.8} (\textbf{81.8})  \\
DeepSeek-8B  & 10.1 & 86.0  & 16.3 & 89.2 (\(\textbf{90.0}\)) & \textbf{9.8} & 86.0 (86.6)   \\
Phi-4-RP-14B & \textbf{6.4} & 77.8  & 8.9 & \textbf{80.0} (\textbf{80.0}) & 6.7 & 77.0 (76.6) \\
\midrule

\rowcolor{paired-light-purple} \Block{1-7}{\textbf{AIME25}} \\
\midrule
Qwen3-8B      & 11.6 & 75.4 & 13.8 & 69.0 (\(71.0\)) & \textbf{9.1} & \textbf{75.6} (74.4) \\
DeepSeek-8B   & 12.0 & 83.3 & 18.9 & 83.3\ (\(\textbf{83.6}\)) & \textbf{11.6} & 82.4 (83.3)  \\
Phi-4-RP-14B  & 13.5 & 62.6 & 11.2 & 76.3 (\textbf{76.3}) & \textbf{8.3} & 64.6 (67.3) \\
\midrule

\rowcolor{paired-light-purple} \Block{1-7}{\textbf{GPQA-Diamond}} \\
\midrule
Qwen3-8B      & 58.5 & 61.7 & 64.2 & 62.8 (62.3) & \textbf{35.0} & 63.1 (\textbf{63.2}) \\
DeepSeek-8B   & 42.1 & 67.1 & 73.8 & \textbf{67.3} (66.3) & \textbf{34.5} & 62.1 (63.1)  \\
Phi-4-RP-14B  & 41.8 & 58.1 & 49.3 & \textbf{72.7} (72.3) & \textbf{37.8} & 70.7 (63.2) \\
\bottomrule
\end{NiceTabular}
  \caption{Accuracy (in \%) and required tokens (in $\times 10^{6}$) for a branching factor $n=32$. For Flex-\method, we set the seed path length $l_0=2,048$ and also include it as an additional cost in the required token computation. We report majority voting (MV) and confidence-weighted majority voting (WMV) scores. Flex-\method integrates Adaptive Self-Consistency to show that Fork-think can be adapted easily with efficiency mechanisms to achieve even higher efficiency gains.}
  \label{tab:add_res}
\end{table}

\section{Analysis}\label{sec:06-results}

We analyze our results in terms of the forking position and nature of forking tokens, and conduct additional ablation studies for different model confidence estimates (Equation~\ref{eq:confidence}) as well as varying branching factors $n$ and seed path length $l_0$.

\begin{figure}[!tb]
     \centering
     \hfill
     \begin{subfigure}[b]{0.48\textwidth}
         \centering
         \includegraphics[width=\textwidth]{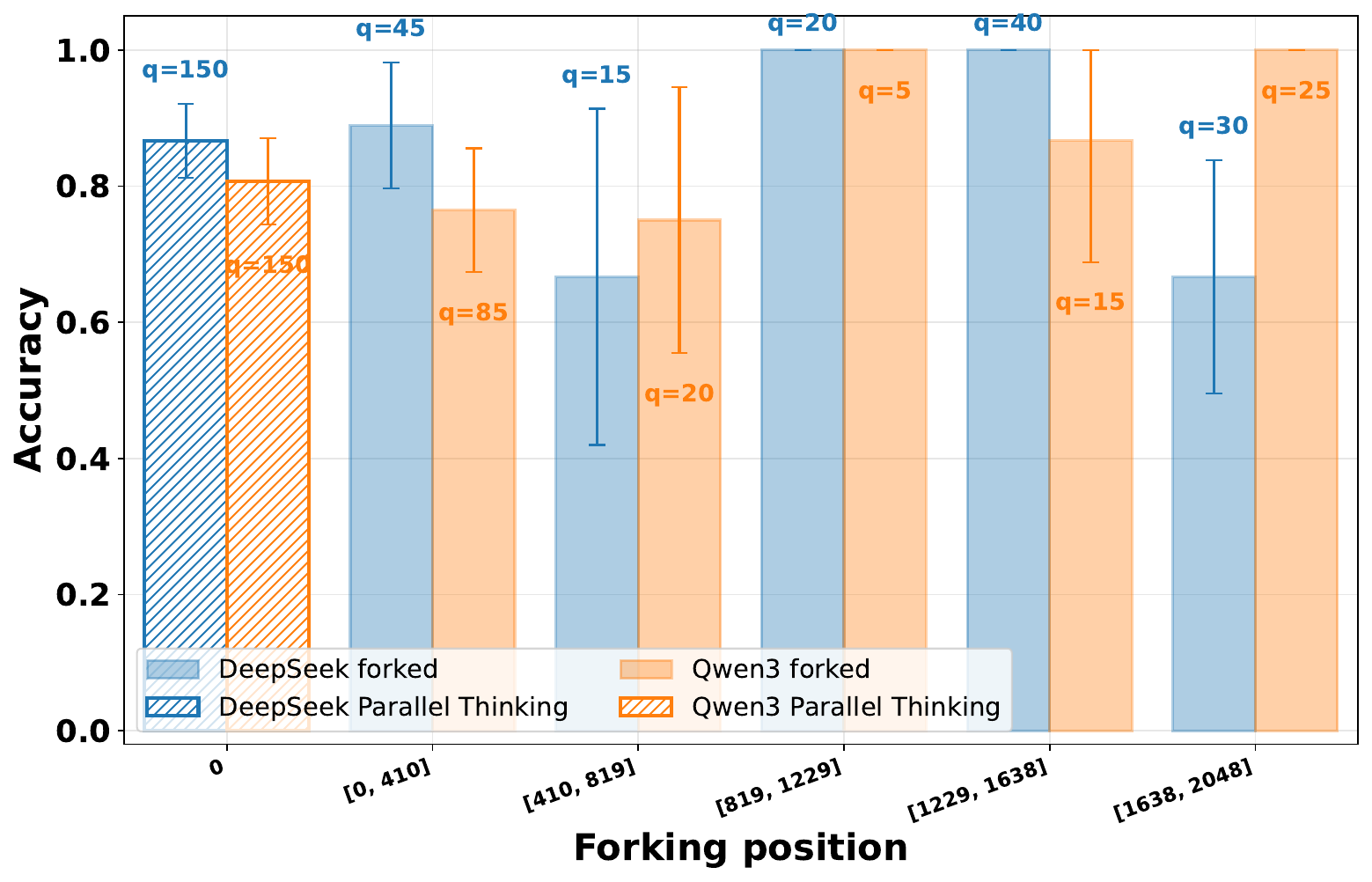}
         \caption{AIME24}
         \label{fig:AIME24-pos}
     \end{subfigure}
     \hfill
     \begin{subfigure}[b]{0.48\textwidth}
         \centering
         \includegraphics[width=\textwidth]{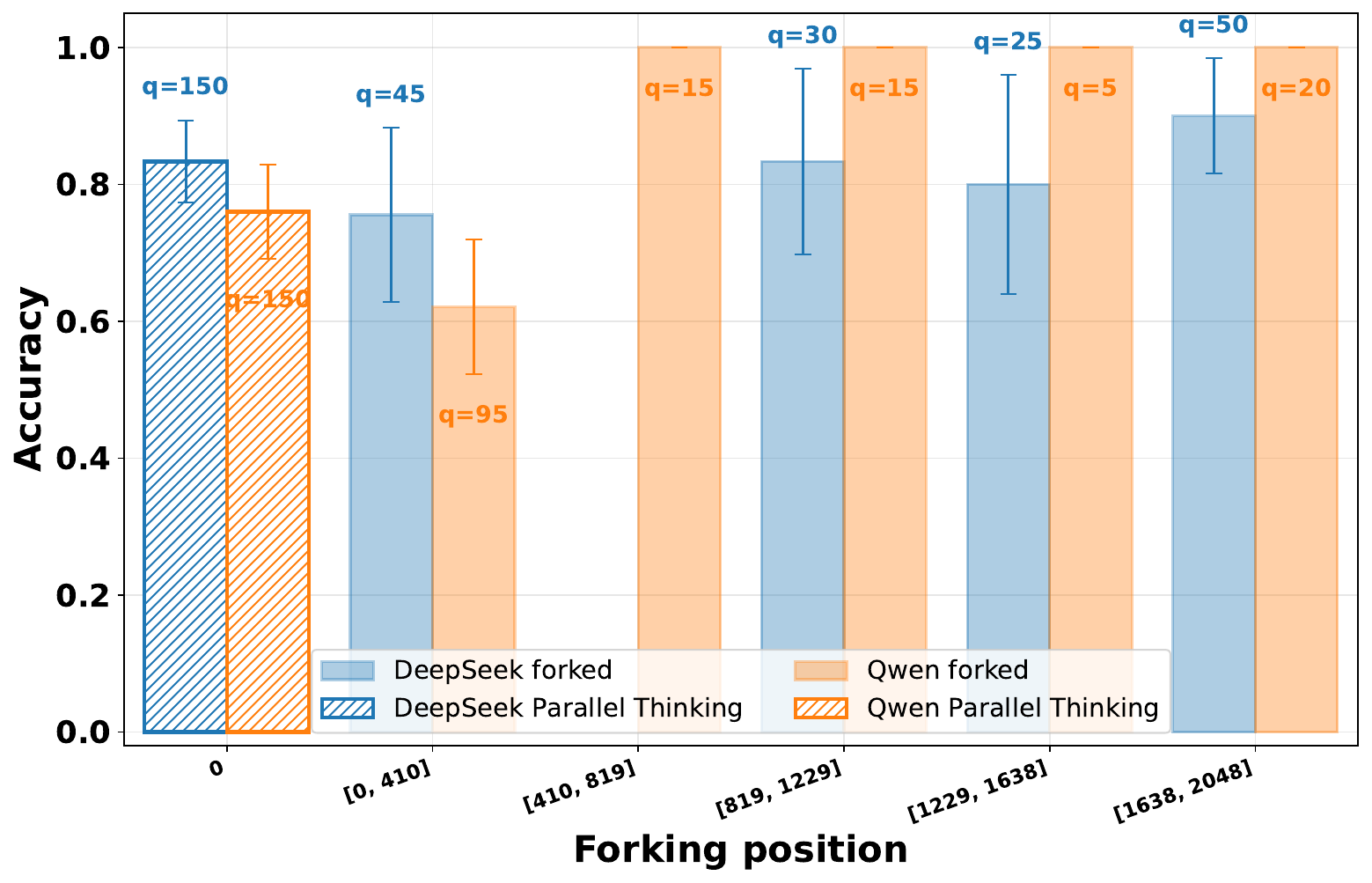}
         \caption{AIME25}
         \label{fig:AIME25-pos}
     \end{subfigure}
        \caption{Accuracy (using majority voting), averaged over each bin of different forking point positions (for \method) with 95\% confidence intervals. Parallel Thinking always forks at position zero. We can see that in multiple instances, forking later in the seed path achieves a substantially higher performance compared to Parallel Thinking, providing evidence that identifying good forking points can substantially boost the performance (while reducing compute cost).}
        \label{fig:fork-pos}
\end{figure}

\paragraph{Forking Position.} 
For each AIME dataset,
we study the effect of the forking position by binning each forking point with respect to its position in the greedy seed path. 
\Cref{fig:fork-pos} shows the accuracy (averaged across each bin) for forking points attributed to one of five equidistant bins starting from 0 to 2,048 ($l_0$). 
In addition, we report the average performance for Parallel Thinking at position~0. 
Interestingly, we observe that some forking points (e.g., all points in the second or later bins for AIME25) exhibit a substantially higher performance than others, indicating the importance of choosing the right forking point.
For AIME24, we can also observe model-specific differences as points from the third and fifth bins perform best for the Qwen3-8B model while for DeepSeek-8B points from the third and fourth bins perform best.
Finally, we find that forking too early (in the first or second bins) often seems to hurt model performance, indicating that the commonly used \emph{think-first-then-decide} paradigm might not always be optimal.

\paragraph{Forking Tokens.} We study the nature of forking tokens induced by \method by analyzing the top-10 most prominent forking tokens across all models on the AIME datasets and find that these tokens correspond to compact mathematical syntax, including variables and operators such as \{a, (x , b , =, -, +, 5, 2, 4, 1\}. While these tokens may look syntactically small, they are functionally important in mathematical reasoning. A choice of equality, sign, operator, variable assignment, or algebraic transformation can determine the subsequent reasoning step and whether the solution remains correct or diverges. This is consistent with prior observations that pivotal or critical tokens in math reasoning do not always correspond to human-salient semantic words, and often include operators or numerical manipulation tokens \citep{abdin2024phi4technicalreport, lin2025critical}. We further analyze the distribution of forking tokens relative to the solution steps in the reasoning trace on AIME24 and find that 83\% of the forking points happen right after or in the middle of a key reasoning step, e.g., in the middle of an equation, while the rest of the points happen near text continuation that are not particularly key reasoning steps. This indicates that the forking points selected by \method are important divergence points for the correctness of the solution.

\subsection{Ablation Experiments}
We conduct ablation experiments for varying the branching factor $n$ as well as different model confidence estimation methods.

\paragraph{Impact of the Branching Factor.}
Many existing methods often require a high branching factor $n$.\footnote{For instance, \citet{fu2026deep} utilize 512 parallel traces in their experiments using an undisclosed compute infrastructure.}
However, using large $n$ is often not possible for resource-constrained, academic setups.  
We thus study the scaling behavior of \method at small $n \in \{4,8,16,32\}$. 
\Cref{{fig:Num-samples}} shows the averaged accuracy (using majority voting) and the number of required tokens (in millions) of \method against Parallel Thinking. 
We can observe that \method consumes substantially fewer tokens across all $n$ and maintains a competitive performance to Parallel Thinking on AIME25 and even outperforms it on AIME24.
We conclude that \method offers a cheap alternative to Parallel Thinking even with a small branching factor. 

\begin{figure}[h]
     \centering
     \hfill
     \begin{subfigure}[b]{0.49\textwidth}
         \centering
         \includegraphics[width=\textwidth]{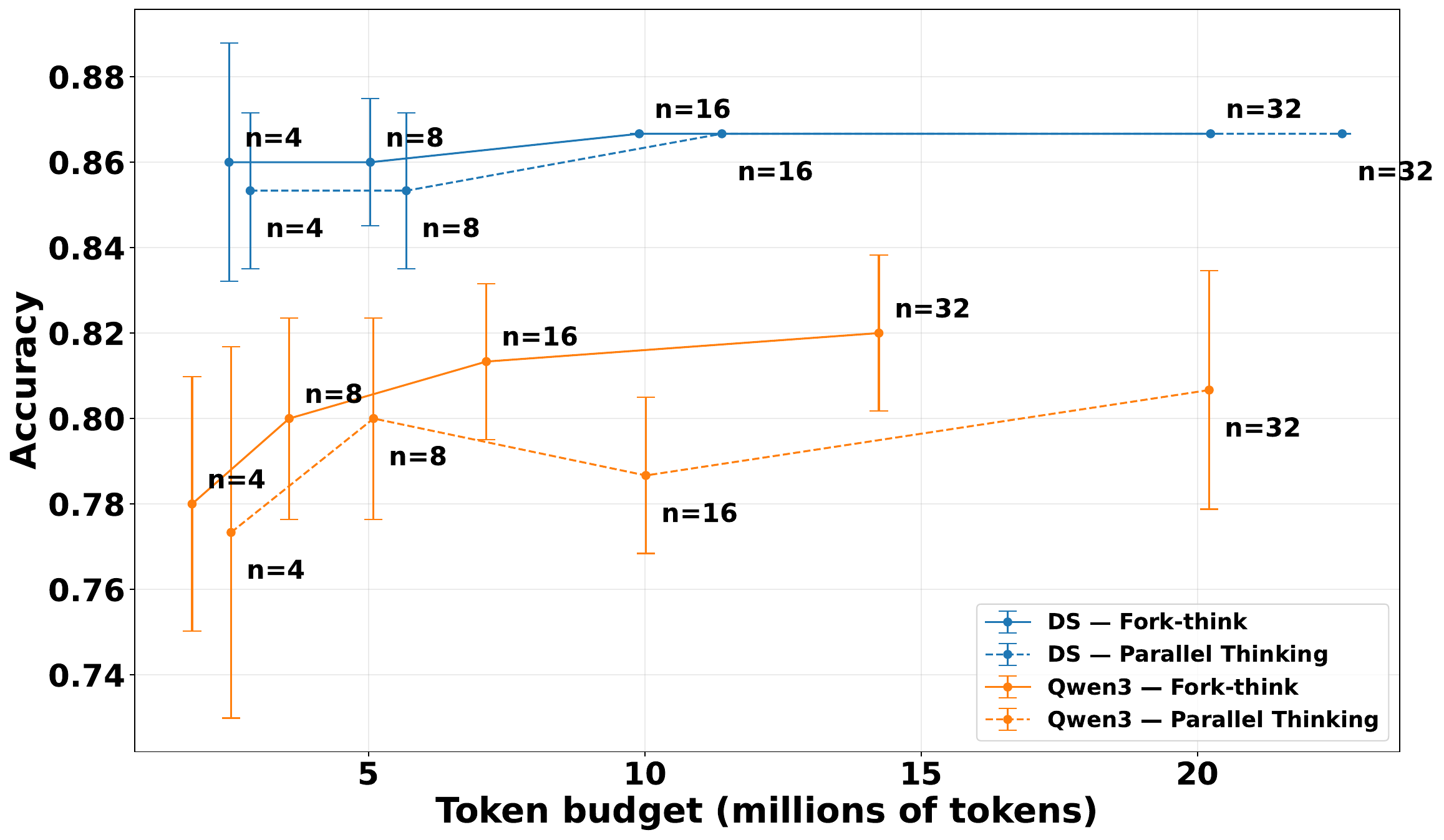}
         \caption{AIME24}
         \label{fig:three sin x}
     \end{subfigure}
     \hfill
     \begin{subfigure}[b]{0.49\textwidth}
         \centering
         \includegraphics[width=\textwidth]{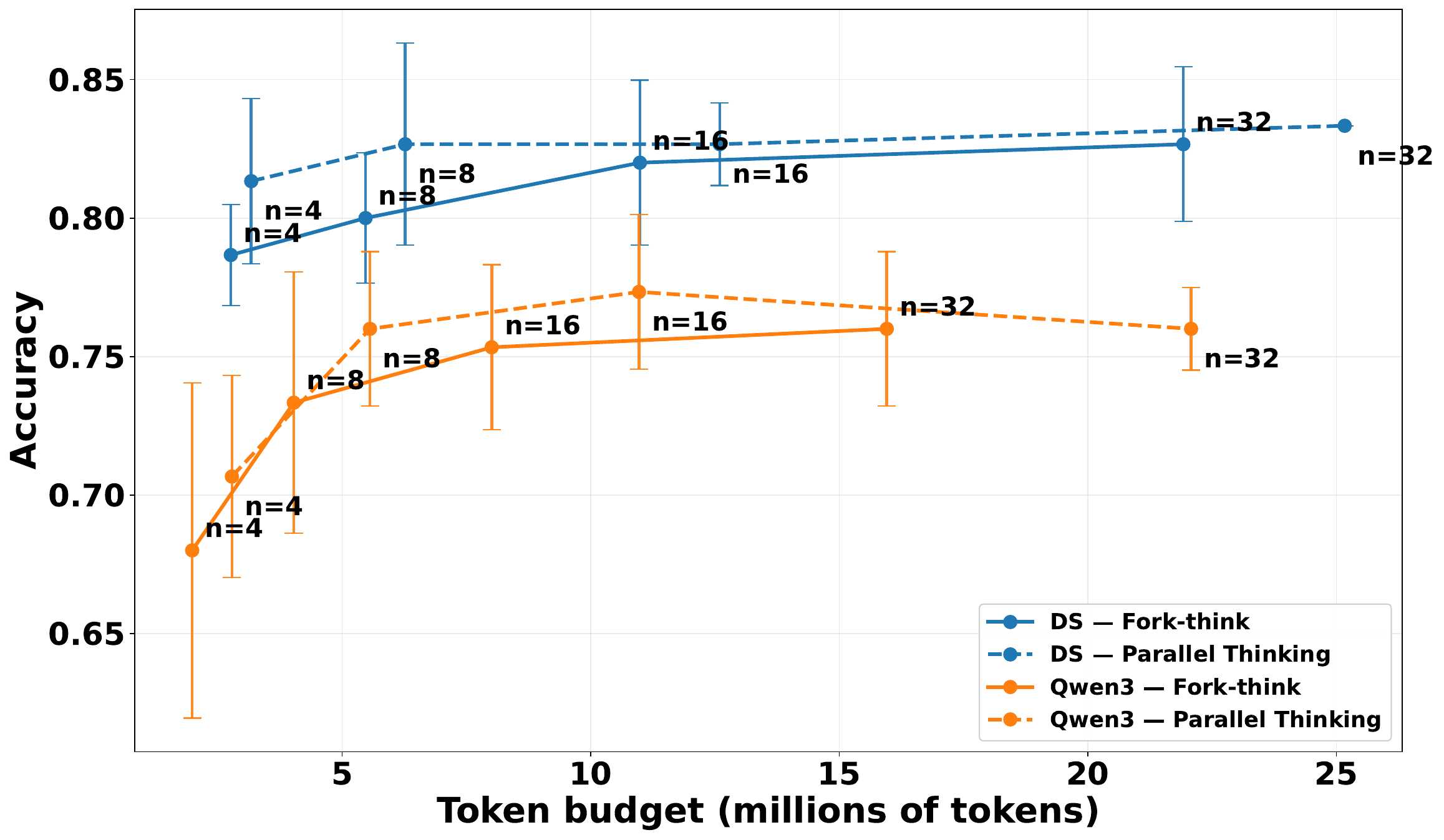}
         \caption{AIME25}
         \label{fig:five over x}
     \end{subfigure}
        \caption{Scaling behavior of Forkthink against baseline Parallel think. Models’ accuracy vs. tokens ($\times 10^{6}$) for different methods on different datasets.}
        \label{fig:Num-samples}
\end{figure}

\paragraph{Impact of the Seed Path Length.} To further understand the trade-off between the length of the seed path $l_0$ and tokens spent in the branches, we study the impact of seed path length on overall token usage as well as performance. \Cref{tab:seed_path} shows the token usage and accuracy of \method with Qwen3-8B on AIME24 when varying $l_0 \in \{512, 1024, 2048, 4096\}$. We can see that Fork-Think is not sensitive with respect to accuracy. This is likely due to most forking tokens being centered earlier rather than later (cf. \Cref{fig:fork-pos}). Hence, increasing $l_0$ does not affect the selected forking token. Regarding token usage, we do find differences, with a too-short $l_0$ causing too-early branching for some of the questions, and a too-long $l_0$ leading to overspending tokens. These results highlight the importance of $l_0$ length on the overall token usage of \method. 
\begin{table}[!htb]
\centering
\small
\begin{tabular}{lrrrr}
\toprule
 $l_0$ Length & 512 & 1024 & 2048 & 4096\\
\midrule
Accuracy (in \%) & 81.1 & 81.1 & \textbf{81.8} & \textbf{81.8}  \\
Tokens (in $\times 10^{6}$) & 16.0 & 15.2 & \textbf{14.2} & 15.2    \\
\bottomrule
\end{tabular}
\caption{\method with respect to different seed path length $l_0$ using Qwen3-8B on AIME24, with branching factor n=32. }
\label{tab:seed_path}
\end{table}

\paragraph{Different Model Confidence Estimates.}
Although we utilize the average log-probability (cf. Equation~\ref{eq:confidence}) in this work as a proxy for model confidence, various other metrics have been proposed in the past.
We hence extend our evaluation to token entropy and select the forking point according to the highest entropy over the normalized top-$k$ probabilities:
\begin{equation}
   H_i = -\sum_{j=1}^{k} \mathrm{P}(x_{j})\log \mathrm{P}(x_{j}),
   \label{eq:entropy}
\end{equation}
where $\mathrm{P}(x_{j}) = \frac{P(x_j)}{\sum_{i=1}^{k} P(x_i)}$ is the normalized probability of the $j$-th token.
We further conduct experiments with different $k$ and evaluate a variant of \method where we select the maximum instead of the minimum average confidence point (cf. \Cref{eq:least-confidence}).

\Cref{tab:forking-metric} shows the averaged accuracy (using majority voting) as well as the overall token consumption for all model confidence estimation metrics. 
We can see that using the model confidence estimate provided in Equation~\ref{eq:confidence} uses fewer tokens overall. 
This result suggests that, although the average log-probability is a heuristic that depends strongly on the shape and sharpness of the top-$k$ distribution at each position, it provides an effective and efficient signal for selecting forking points in our setting. 
In contrast, entropy over the normalized top-$k$ distribution might be a more principled measure of uncertainty, but it does not lead to better performance in our experiments. 
We provide the results for all models and both AIME datasets in appendix~\ref{sec:appendix-FM}.
In future work, we will investigate more sophisticated model confidence estimation methods---for instance, ones based on previous decoding works such as locally typical sampling~\citep{meister_locally_2023}.

\begin{table}[!htb]
\centering
\small
\begin{tabular}{lrrrrr}
\toprule
 \textbf{Model Confidence} & $\arg\min_1$ & Entropy$_{10}$ & Entropy$_{1000}$ & $\arg\min_{10}$ & $\arg\max_{10}$ \\
\midrule
Accuracy (in \%)  & 82.6  & \textbf{86.0} & \textbf{86.0} & \textbf{86.0} & \textbf{86.0}  \\
Tokens (in $\times 10^{6}$) & 20.15 & 21.1 & 21.1 & \textbf{20.14} & 21.1  \\
\bottomrule
\end{tabular}
\caption{Averaged accuracy (majority voting) for different model confidence estimates using the DeepSeek-8B model on AIME24. We set the number of branching factor $n=32$.}
\label{tab:forking-metric}
\end{table}

\section{Conclusion}\label{sec:07-conclusion}

We proposed \method, a method that shifts the perspective of Parallel Thinking from the predominantly used \emph{think-first-then-decide} paradigm to a \emph{decide-first-then-think} paradigm.
In contrast to prior methods that by default sample parallel traces from the very beginning, \method utilizes a greedy seed path to first identify forking points for sampling.

Our experiments across three benchmarks using three reasoning models show that \method achieves a performance comparable to Parallel Thinking (both using majority voting) but using 7--30\% less tokens and requiring 38--57\% less run-time.
The results further show that a variant of \method, combined with early stopping techniques, is also competitive against other baselines, including a recent state-of-the-art method that utilizes adaptive pruning together with confidence-weighted majority voting.
Our analysis reveals that using early forking points by default might not be a good choice, as later forking points often lead to substantially better parallel traces. 
We further show that \method's performance remains competitive to Parallel Thinking even in low-budget scenarios that use a small branching factor (4, 8, 16). 

These results suggest multiple promising directions for future research. 
First, the high performance of a single forking point suggests that increasing the number of forking points (i.e., sampling a smaller number of parallel traces at multiple locations) could be a promising direction.
Second, the dependence of token saving on the seed path $l_0$ length motivates the need for an adaptive or budget-aware seed path that dynamically adjusts the length based on the problem at hand.
Finally, we will explore other methods to estimate the model confidence (beyond entropy) which could substantially improve our understanding on selecting forking points.

\section*{Acknowledgments}
We thank Michael Hahn, Dawei Zhu, and Sergey Troshin for their helpful discussions and feedback on this project. We are also grateful to anonymous reviewers for their helpful comments. 
This work was funded by 
the Deutsche Forschungsgemeinschaft (DFG, German Research Foundation) -- SFB 1102 under project 232722074, 
and GRK 2853/1 “Neuroexplicit Models of Language, Vision, and Action” - under project 471607914.

\section*{Ethics Statement}

\paragraph{Disclosure of LLM Use.}
Generative AI tools were used solely to assist with debugging code, polish the manuscript (via paraphrasing), and for fixing grammar and typographical issues. 
These tools were not used to produce any scientific content, including the experimental design, analysis, citations, or conclusions. 
All authors have reviewed and approved the manuscript and assume full responsibility for its contents.

\bibliography{colm2026_conference}

@inproceedings{
kang2025scalable,
title={Scalable Best-of-N Selection for Large Language Models via Self-Certainty},
author={Zhewei Kang and Xuandong Zhao and Dawn Song},
booktitle={The Thirty-ninth Annual Conference on Neural Information Processing Systems},
year={2025},
url={https://openreview.net/forum?id=29FRqmVQK8}
}

@inproceedings{
xie2022an,
title={An Explanation of In-context Learning as Implicit Bayesian Inference},
author={Sang Michael Xie and Aditi Raghunathan and Percy Liang and Tengyu Ma},
booktitle={International Conference on Learning Representations},
year={2022},
url={https://openreview.net/forum?id=RdJVFCHjUMI}
}

@misc{zheng2024sglangefficientexecutionstructured,
      title={SGLang: Efficient Execution of Structured Language Model Programs}, 
      author={Lianmin Zheng and Liangsheng Yin and Zhiqiang Xie and Chuyue Sun and Jeff Huang and Cody Hao Yu and Shiyi Cao and Christos Kozyrakis and Ion Stoica and Joseph E. Gonzalez and Clark Barrett and Ying Sheng},
      year={2024},
      eprint={2312.07104},
      archivePrefix={arXiv},
      primaryClass={cs.AI},
      url={https://arxiv.org/abs/2312.07104}, 
}

@inproceedings{kwon2023vllm,
  title={Efficient Memory Management for Large Language Model Serving with PagedAttention},
  author={Woosuk Kwon and Zhuohan Li and Siyuan Zhuang and Ying Sheng and Lianmin Zheng and Cody Hao Yu and Joseph E. Gonzalez and Hao Zhang and Ion Stoica},
  booktitle={Proceedings of the ACM SIGOPS 29th Symposium on Operating Systems Principles},
  year={2023}
}

@inproceedings{
dang2025weight,
title={Weight ensembling improves reasoning in language models},
author={Xingyu Dang and Christina Baek and Kaiyue Wen and J Zico Kolter and Aditi Raghunathan},
booktitle={Second Conference on Language Modeling},
year={2025},
url={https://openreview.net/forum?id=S2IKxulLT1}
}

@inproceedings{he-etal-2025-rewarding,
    title = "Rewarding the Unlikely: Lifting {GRPO} Beyond Distribution Sharpening",
    author = "He, Andre Wang  and
      Fried, Daniel  and
      Welleck, Sean",
    editor = "Christodoulopoulos, Christos  and
      Chakraborty, Tanmoy  and
      Rose, Carolyn  and
      Peng, Violet",
    booktitle = "Proceedings of the 2025 Conference on Empirical Methods in Natural Language Processing",
    month = nov,
    year = "2025",
    address = "Suzhou, China",
    publisher = "Association for Computational Linguistics",
    url = "https://aclanthology.org/2025.emnlp-main.1298/",
    doi = "10.18653/v1/2025.emnlp-main.1298",
    pages = "25548--25560",
    ISBN = "979-8-89176-332-6",
}

@misc{karan2025reasoningsamplingbasemodel,
      title={Reasoning with Sampling: Your Base Model is Smarter Than You Think}, 
      author={Aayush Karan and Yilun Du},
      year={2025},
      eprint={2510.14901},
      archivePrefix={arXiv},
      primaryClass={cs.LG},
      url={https://arxiv.org/abs/2510.14901}, 
}

@inproceedings{li-etal-2023-making,
    title = "Making Language Models Better Reasoners with Step-Aware Verifier",
    author = "Li, Yifei  and
      Lin, Zeqi  and
      Zhang, Shizhuo  and
      Fu, Qiang  and
      Chen, Bei  and
      Lou, Jian-Guang  and
      Chen, Weizhu",
    editor = "Rogers, Anna  and
      Boyd-Graber, Jordan  and
      Okazaki, Naoaki",
    booktitle = "Proceedings of the 61st Annual Meeting of the Association for Computational Linguistics (Volume 1: Long Papers)",
    month = jul,
    year = "2023",
    address = "Toronto, Canada",
    publisher = "Association for Computational Linguistics",
    url = "https://aclanthology.org/2023.acl-long.291/",
    doi = "10.18653/v1/2023.acl-long.291",
    pages = "5315--5333",
}

@inproceedings{
song2025outcomebased,
title={Outcome-based Exploration for {LLM} Reasoning},
author={Yuda Song and Julia Kempe and R{\'e}mi Munos},
booktitle={NeurIPS 2025 Workshop: Second Workshop on Aligning Reinforcement Learning Experimentalists and Theorists},
year={2025},
url={https://openreview.net/forum?id=VORSpYLBJ6}
}

@article{yao2023tree,
  title={Tree of thoughts: Deliberate problem solving with large language models},
  author={Yao, Shunyu and Yu, Dian and Zhao, Jeffrey and Shafran, Izhak and Griffiths, Tom and Cao, Yuan and Narasimhan, Karthik},
  journal={Advances in neural information processing systems},
  volume={36},
  pages={11809--11822},
  year={2023}
}

@InProceedings{zhao2025scalinglaw,
  title = 	 {Sample, Scrutinize and Scale: Effective Inference-Time Search by Scaling Verification},
  author =       {Zhao, Eric and Awasthi, Pranjal and Gollapudi, Sreenivas},
  booktitle = 	 {Proceedings of the 42nd International Conference on Machine Learning},
  pages = 	 {77272--77309},
  year = 	 {2025},
  editor = 	 {Singh, Aarti and Fazel, Maryam and Hsu, Daniel and Lacoste-Julien, Simon and Berkenkamp, Felix and Maharaj, Tegan and Wagstaff, Kiri and Zhu, Jerry},
  volume = 	 {267},
  series = 	 {Proceedings of Machine Learning Research},
  month = 	 {13--19 Jul},
  publisher =    {PMLR},
  url = 	 {https://proceedings.mlr.press/v267/zhao25a.html},
}

@inproceedings{
inoue2025wider,
title={Wider or Deeper?  Scaling {LLM} Inference-Time Compute with Adaptive Branching Tree Search},
author={Yuichi Inoue and Kou Misaki and Yuki Imajuku and So Kuroki and Taishi Nakamura and Takuya Akiba},
booktitle={The Thirty-ninth Annual Conference on Neural Information Processing Systems},
year={2025},
url={https://openreview.net/forum?id=jAsr5GHt3P}
}

@article{
welleck2024from,
title={From Decoding to Meta-Generation: Inference-time Algorithms for Large Language Models},
author={Sean Welleck and Amanda Bertsch and Matthew Finlayson and Hailey Schoelkopf and Alex Xie and Graham Neubig and Ilia Kulikov and Zaid Harchaoui},
journal={Transactions on Machine Learning Research},
issn={2835-8856},
year={2024},
url={https://openreview.net/forum?id=eskQMcIbMS},
note={Survey Certification}
}

@inproceedings{
setlur2025rewarding,
title={Rewarding Progress: Scaling Automated Process Verifiers for {LLM} Reasoning},
author={Amrith Setlur and Chirag Nagpal and Adam Fisch and Xinyang Geng and Jacob Eisenstein and Rishabh Agarwal and Alekh Agarwal and Jonathan Berant and Aviral Kumar},
booktitle={The Thirteenth International Conference on Learning Representations},
year={2025},
url={https://openreview.net/forum?id=A6Y7AqlzLW}
}

@inproceedings{
feng2026vericot,
title={VeriCoT: Neuro-symbolic Chain-of-Thought Validation via Logical Consistency Checks},
author={Yu Feng and Nathaniel Weir and Kaj Bostrom and Sam Bayless and Darion Cassel and Sapana Chaudhary and Benjamin Kiesl-Reiter and Huzefa Rangwala},
booktitle={The Fourteenth International Conference on Learning Representations},
year={2026},
url={https://openreview.net/forum?id=zHuV3Vatov}
}

@inproceedings{
ni2023learning,
title={Learning Math Reasoning from Self-Sampled Correct and Partially-Correct Solutions},
author={Ansong Ni and Jeevana Priya Inala and Chenglong Wang and Alex Polozov and Christopher Meek and Dragomir Radev and Jianfeng Gao},
booktitle={The Eleventh International Conference on Learning Representations },
year={2023},
url={https://openreview.net/forum?id=4D4TSJE6-K}
}

@inproceedings{
hosseini2024vstar,
title={V-{ST}aR: Training Verifiers for Self-Taught Reasoners},
author={Arian Hosseini and Xingdi Yuan and Nikolay Malkin and Aaron Courville and Alessandro Sordoni and Rishabh Agarwal},
booktitle={First Conference on Language Modeling},
year={2024},
url={https://openreview.net/forum?id=stmqBSW2dV}
}

@article{guo2025deepseek,
  title={DeepSeek-R1 incentivizes reasoning in LLMs through reinforcement learning},
  author={Guo, Daya and Yang, Dejian and Zhang, Haowei and Song, Junxiao and Wang, Peiyi and Zhu, Qihao and Xu, Runxin and Zhang, Ruoyu and Ma, Shirong and Bi, Xiao and others},
  journal={Nature},
  volume={645},
  number={8081},
  pages={633--638},
  year={2025},
  publisher={Nature Publishing Group UK London}
}

@misc{wang2024openropensourceframework,
      title={OpenR: An Open Source Framework for Advanced Reasoning with Large Language Models}, 
      author={Jun Wang and Meng Fang and Ziyu Wan and Muning Wen and Jiachen Zhu and Anjie Liu and Ziqin Gong and Yan Song and Lei Chen and Lionel M. Ni and Linyi Yang and Ying Wen and Weinan Zhang},
      year={2024},
      eprint={2410.09671},
      archivePrefix={arXiv},
      primaryClass={cs.AI},
      url={https://arxiv.org/abs/2410.09671}, 
}

@inproceedings{wei2022cot,
author = {Wei, Jason and Wang, Xuezhi and Schuurmans, Dale and Bosma, Maarten and Ichter, Brian and Xia, Fei and Chi, Ed H. and Le, Quoc V. and Zhou, Denny},
title = {Chain-of-thought prompting elicits reasoning in large language models},
year = {2022},
isbn = {9781713871088},
publisher = {Curran Associates Inc.},
address = {Red Hook, NY, USA},
booktitle = {Proceedings of the 36th International Conference on Neural Information Processing Systems},
articleno = {1800},
numpages = {14},
location = {New Orleans, LA, USA},
series = {NIPS '22}
}

@inproceedings{takeshi2022zeroshot,
author = {Kojima, Takeshi and Gu, Shixiang Shane and Reid, Machel and Matsuo, Yutaka and Iwasawa, Yusuke},
title = {Large language models are zero-shot reasoners},
year = {2022},
isbn = {9781713871088},
publisher = {Curran Associates Inc.},
address = {Red Hook, NY, USA},
booktitle = {Proceedings of the 36th International Conference on Neural Information Processing Systems},
articleno = {1613},
numpages = {15},
location = {New Orleans, LA, USA},
series = {NIPS '22}
}

@misc{kumar2025llmposttrainingdeepdive,
      title={LLM Post-Training: A Deep Dive into Reasoning Large Language Models}, 
      author={Komal Kumar and Tajamul Ashraf and Omkar Thawakar and Rao Muhammad Anwer and Hisham Cholakkal and Mubarak Shah and Ming-Hsuan Yang and Phillip H. S. Torr and Fahad Shahbaz Khan and Salman Khan},
      year={2025},
      eprint={2502.21321},
      archivePrefix={arXiv},
      primaryClass={cs.CL},
      url={https://arxiv.org/abs/2502.21321}, 
}

@inproceedings{rafailov2023dpo,
 author = {Rafailov, Rafael and Sharma, Archit and Mitchell, Eric and Manning, Christopher D and Ermon, Stefano and Finn, Chelsea},
 booktitle = {Advances in Neural Information Processing Systems},
 editor = {A. Oh and T. Naumann and A. Globerson and K. Saenko and M. Hardt and S. Levine},
 pages = {53728--53741},
 publisher = {Curran Associates, Inc.},
 title = {Direct Preference Optimization: Your Language Model is Secretly a Reward Model},
 url = {https://proceedings.neurips.cc/paper_files/paper/2023/file/a85b405ed65c6477a4fe8302b5e06ce7-Paper-Conference.pdf},
 volume = {36},
 year = {2023}
}

@article{chung2024scaling,
author = {Chung, Hyung Won and Hou, Le and Longpre, Shayne and Zoph, Barret and Tai, Yi and Fedus, William and Li, Yunxuan and Wang, Xuezhi and Dehghani, Mostafa and Brahma, Siddhartha and Webson, Albert and Gu, Shixiang Shane and Dai, Zhuyun and Suzgun, Mirac and Chen, Xinyun and Chowdhery, Aakanksha and Castro-Ros, Alex and Pellat, Marie and Robinson, Kevin and Valter, Dasha and Narang, Sharan and Mishra, Gaurav and Yu, Adams and Zhao, Vincent and Huang, Yanping and Dai, Andrew and Yu, Hongkun and Petrov, Slav and Chi, Ed H. and Dean, Jeff and Devlin, Jacob and Roberts, Adam and Zhou, Denny and Le, Quoc V. and Wei, Jason},
title = {Scaling instruction-finetuned language models},
year = {2024},
issue_date = {January 2024},
publisher = {JMLR.org},
volume = {25},
number = {1},
issn = {1532-4435},
journal = {J. Mach. Learn. Res.},
month = jan,
articleno = {70},
numpages = {53}
}

@inproceedings{
wang2023selfconsistency,
title={Self-Consistency Improves Chain of Thought Reasoning in Language Models},
author={Xuezhi Wang and Jason Wei and Dale Schuurmans and Quoc V Le and Ed H. Chi and Sharan Narang and Aakanksha Chowdhery and Denny Zhou},
booktitle={The Eleventh International Conference on Learning Representations },
year={2023},
url={https://openreview.net/forum?id=1PL1NIMMrw}
}

@inproceedings{
wang2024mmlupro,
title={{MMLU}-Pro: A More Robust and Challenging Multi-Task Language Understanding Benchmark},
author={Yubo Wang and Xueguang Ma and Ge Zhang and Yuansheng Ni and Abhranil Chandra and Shiguang Guo and Weiming Ren and Aaran Arulraj and Xuan He and Ziyan Jiang and Tianle Li and Max Ku and Kai Wang and Alex Zhuang and Rongqi Fan and Xiang Yue and Wenhu Chen},
booktitle={The Thirty-eight Conference on Neural Information Processing Systems Datasets and Benchmarks Track},
year={2024},
url={https://openreview.net/forum?id=y10DM6R2r3}
}

@inproceedings{
wu2025inference,
title={Inference Scaling Laws: An Empirical Analysis of Compute-Optimal Inference for {LLM} Problem-Solving},
author={Yangzhen Wu and Zhiqing Sun and Shanda Li and Sean Welleck and Yiming Yang},
booktitle={The Thirteenth International Conference on Learning Representations},
year={2025},
url={https://openreview.net/forum?id=VNckp7JEHn}
}

@article{balachandran2025inference,
  title={Inference-time scaling for complex tasks: Where we stand and what lies ahead},
  author={Balachandran, Vidhisha and Chen, Jingya and Chen, Lingjiao and Garg, Shivam and Joshi, Neel and Lara, Yash and Langford, John and Nushi, Besmira and Vineet, Vibhav and Wu, Yue and others},
  journal={arXiv preprint arXiv:2504.00294},
  year={2025}
}

@misc{abdin2024phi4technicalreport,
      title={Phi-4 Technical Report}, 
      author={Marah Abdin and Jyoti Aneja and Harkirat Behl and Sébastien Bubeck and Ronen Eldan and Suriya Gunasekar and Michael Harrison and Russell J. Hewett and Mojan Javaheripi and Piero Kauffmann and James R. Lee and Yin Tat Lee and Yuanzhi Li and Weishung Liu and Caio C. T. Mendes and Anh Nguyen and Eric Price and Gustavo de Rosa and Olli Saarikivi and Adil Salim and Shital Shah and Xin Wang and Rachel Ward and Yue Wu and Dingli Yu and Cyril Zhang and Yi Zhang},
      year={2024},
      eprint={2412.08905},
      archivePrefix={arXiv},
      primaryClass={cs.CL},
      url={https://arxiv.org/abs/2412.08905}, 
}

@inproceedings{
fu2026deep,
title={Deep Think with Confidence},
author={Yichao Fu and Xuewei Wang and Hao Zhang and Yuandong Tian and Jiawei Zhao},
booktitle={The Fourteenth International Conference on Learning Representations},
year={2026},
url={https://openreview.net/forum?id=8LqHs0KIM7}
}

@misc{
leang2026picsar,
title={Pi{CSAR}: Probabilistic Confidence Selection And Ranking for Reasoning Chains},
author={Joshua Ong Jun Leang and Zheng Zhao and Aryo Pradipta Gema and Sohee Yang and Wai-Chung Kwan and Xuanli He and Wenda Li and Pasquale Minervini and Eleonora Giunchiglia and Shay B Cohen},
year={2026},
url={https://openreview.net/forum?id=0nmGpN7Qqa}
}

@inproceedings{
wang2025beyond,
title={Beyond the 80/20 Rule: High-Entropy Minority Tokens Drive Effective Reinforcement Learning for {LLM} Reasoning},
author={Shenzhi Wang and Le Yu and Chang Gao and Chujie Zheng and Shixuan Liu and Rui Lu and Kai Dang and Xiong-Hui Chen and Jianxin Yang and Zhenru Zhang and Yuqiong Liu and An Yang and Andrew Zhao and Yang Yue and Shiji Song and Bowen Yu and Gao Huang and Junyang Lin},
booktitle={The Thirty-ninth Annual Conference on Neural Information Processing Systems},
year={2025},
url={https://openreview.net/forum?id=yfcpdY4gMP}
}

@inproceedings{
troshin2025control,
title={Control the Temperature: Selective Sampling for Diverse and High-Quality {LLM} Outputs},
author={Sergey Troshin and Wafaa Mohammed and Yan Meng and Christof Monz and Antske Fokkens and Vlad Niculae},
booktitle={Second Conference on Language Modeling},
year={2025},
url={https://openreview.net/forum?id=IyOC5GCzv4}
}

@inproceedings{
tao2025revisiting,
title={Revisiting Uncertainty Estimation and Calibration of Large Language Models},
author={Linwei Tao and Yi-Fan Yeh and Minjing Dong and Tao Huang and Jialin Yu and Philip Torr and Chang Xu},
booktitle={Workshop on Scaling Environments for Agents},
year={2025},
url={https://openreview.net/forum?id=Q9CreVjHH7}
}

@misc{balunovic_srimatharena_2025,
  title = {MathArena: Evaluating LLMs on Uncontaminated Math Competitions},
  author = {Mislav Balunović and Jasper Dekoninck and Ivo Petrov and Nikola Jovanović and Martin Vechev},
  copyright = {MIT},
  url = {https://matharena.ai/},
  publisher = {SRI Lab, ETH Zurich},
  month = feb,
  year = {2025},
}

@inproceedings{
lightman2024lets,
title={Let's Verify Step by Step},
author={Hunter Lightman and Vineet Kosaraju and Yuri Burda and Harrison Edwards and Bowen Baker and Teddy Lee and Jan Leike and John Schulman and Ilya Sutskever and Karl Cobbe},
booktitle={The Twelfth International Conference on Learning Representations},
year={2024},
url={https://openreview.net/forum?id=v8L0pN6EOi}
}

@article{hendrycksmath2021,
  title={Measuring Mathematical Problem Solving With the MATH Dataset},
  author={Dan Hendrycks and Collin Burns and Saurav Kadavath and Akul Arora and Steven Basart and Eric Tang and Dawn Song and Jacob Steinhardt},
  journal={NeurIPS},
  year={2021}
}

@misc{AIME24I,
  title = "AIME24 I",
  author = {{AIME24}},
  url = {https://artofproblemsolving.com/wiki/index.php/
2024_AIME_I},
  year=2024,
  note = {Accessed: 2026}
}

@misc{AIME24II,
  title = "AIME24 II",
  author = {{AIME24}},
  year=2024,
  url = {https://artofproblemsolving.com/wiki/index.php/
2024_AIME_II},
  note = {Accessed: 2026}
}

@misc{AIME25I,
  title = "AIME25 I",
  author = {{AIME25}},
  year=2025,
  url = {https://artofproblemsolving.com/wiki/index.php/
2025_AIME_I},
  note = {Accessed: 2026}
}

@misc{AIME25II,
  title = "AIME25 II",
  author = {{AIME25}},
  year=2025,
  url = {https://artofproblemsolving.com/wiki/index.php/
2025_AIME_II},
  note = {Accessed: 2026}
}

@misc{
scalena2026eager,
title={{EAGER}: Entropy-Aware {GE}neRation for Adaptive Inference-Time Scaling},
author={Daniel Scalena and Leonidas Zotos and Elisabetta Fersini and Malvina Nissim and Ahmet {\"U}st{\"u}n},
year={2026},
url={https://openreview.net/forum?id=NRO8xMzCVm}
}

@misc{wu2026efficiencyadaptivitydeeperlook,
      title={From Efficiency to Adaptivity: A Deeper Look at Adaptive Reasoning in Large Language Models}, 
      author={Chao Wu and Baoheng Li and Mingchen Gao and Yu Tian and Zhenyi Wang},
      year={2026},
      eprint={2511.10788},
      archivePrefix={arXiv},
      primaryClass={cs.AI},
      url={https://arxiv.org/abs/2511.10788}, 
}

@inproceedings{
li2024escape,
title={Escape Sky-high Cost: Early-stopping Self-Consistency for Multi-step Reasoning},
author={Yiwei Li and Peiwen Yuan and Shaoxiong Feng and Boyuan Pan and Xinglin Wang and Bin Sun and Heda Wang and Kan Li},
booktitle={The Twelfth International Conference on Learning Representations},
year={2024},
url={https://openreview.net/forum?id=ndR8Ytrzhh}
}

@misc{liang2026hiddenstatesearlysignals,
      title={Hidden States as Early Signals: Step-level Trace Evaluation and Pruning for Efficient Test-Time Scaling}, 
      author={Zhixiang Liang and Beichen Huang and Zheng Wang and Minjia Zhang},
      year={2026},
      eprint={2601.09093},
      archivePrefix={arXiv},
      primaryClass={cs.LG},
      url={https://arxiv.org/abs/2601.09093}, 
}

@misc{tu2025deeppruneparallelscalingintertrace,
      title={DeepPrune: Parallel Scaling without Inter-trace Redundancy}, 
      author={Shangqing Tu and Yaxuan Li and Yushi Bai and Lei Hou and Juanzi Li},
      year={2025},
      eprint={2510.08483},
      archivePrefix={arXiv},
      primaryClass={cs.CL},
      url={https://arxiv.org/abs/2510.08483}, 
}

@inproceedings{taubenfeld-etal-2025-confidence,
    title = "Confidence Improves Self-Consistency in {LLM}s",
    author = "Taubenfeld, Amir  and
      Sheffer, Tom  and
      Ofek, Eran  and
      Feder, Amir  and
      Goldstein, Ariel  and
      Gekhman, Zorik  and
      Yona, Gal",
    editor = "Che, Wanxiang  and
      Nabende, Joyce  and
      Shutova, Ekaterina  and
      Pilehvar, Mohammad Taher",
    booktitle = "Findings of the Association for Computational Linguistics: ACL 2025",
    month = jul,
    year = "2025",
    address = "Vienna, Austria",
    publisher = "Association for Computational Linguistics",
    url = "https://aclanthology.org/2025.findings-acl.1030/",
    doi = "10.18653/v1/2025.findings-acl.1030",
    pages = "20090--20111",
    ISBN = "979-8-89176-256-5",
}

@article{meister_locally_2023,
	title = {Locally {Typical} {Sampling}},
	volume = {11},
	url = {https://aclanthology.org/2023.tacl-1.7/},
	doi = {10.1162/tacl_a_00536},
	urldate = {2025-04-09},
	journal = {Transactions of the Association for Computational Linguistics},
	publisher = {MIT Press},
	author = {Meister, Clara and Pimentel, Tiago and Wiher, Gian and Cotterell, Ryan},
	year = {2023},
	note = {Place: Cambridge, MA},
	pages = {102--121},
}

@inproceedings{wang-etal-2025-ranked,
    title = "Ranked Voting based Self-Consistency of Large Language Models",
    author = "Wang, Weiqin  and
      Wang, Yile  and
      Huang, Hui",
    editor = "Che, Wanxiang  and
      Nabende, Joyce  and
      Shutova, Ekaterina  and
      Pilehvar, Mohammad Taher",
    booktitle = "Findings of the Association for Computational Linguistics: ACL 2025",
    month = jul,
    year = "2025",
    address = "Vienna, Austria",
    publisher = "Association for Computational Linguistics",
    url = "https://aclanthology.org/2025.findings-acl.744/",
    doi = "10.18653/v1/2025.findings-acl.744",
    pages = "14410--14426",
    ISBN = "979-8-89176-256-5",
}

@misc{yang2025qwen3technicalreport,
      title={Qwen3 Technical Report}, 
      author={An Yang and Anfeng Li and Baosong Yang and Beichen Zhang and Binyuan Hui and Bo Zheng and Bowen Yu and Chang Gao and Chengen Huang and Chenxu Lv and Chujie Zheng and Dayiheng Liu and Fan Zhou and Fei Huang and Feng Hu and Hao Ge and Haoran Wei and Huan Lin and Jialong Tang and Jian Yang and Jianhong Tu and Jianwei Zhang and Jianxin Yang and Jiaxi Yang and Jing Zhou and Jingren Zhou and Junyang Lin and Kai Dang and Keqin Bao and Kexin Yang and Le Yu and Lianghao Deng and Mei Li and Mingfeng Xue and Mingze Li and Pei Zhang and Peng Wang and Qin Zhu and Rui Men and Ruize Gao and Shixuan Liu and Shuang Luo and Tianhao Li and Tianyi Tang and Wenbiao Yin and Xingzhang Ren and Xinyu Wang and Xinyu Zhang and Xuancheng Ren and Yang Fan and Yang Su and Yichang Zhang and Yinger Zhang and Yu Wan and Yuqiong Liu and Zekun Wang and Zeyu Cui and Zhenru Zhang and Zhipeng Zhou and Zihan Qiu},
      year={2025},
      eprint={2505.09388},
      archivePrefix={arXiv},
      primaryClass={cs.CL},
      url={https://arxiv.org/abs/2505.09388}, 
}

@inproceedings{wang-etal-2023-chatgpt-defend,
    title = "Can {C}hat{GPT} Defend its Belief in Truth? Evaluating {LLM} Reasoning via Debate",
    author = "Wang, Boshi  and
      Yue, Xiang  and
      Sun, Huan",
    editor = "Bouamor, Houda  and
      Pino, Juan  and
      Bali, Kalika",
    booktitle = "Findings of the Association for Computational Linguistics: EMNLP 2023",
    month = dec,
    year = "2023",
    address = "Singapore",
    publisher = "Association for Computational Linguistics",
    url = "https://aclanthology.org/2023.findings-emnlp.795/",
    doi = "10.18653/v1/2023.findings-emnlp.795",
    pages = "11865--11881",
}

@inproceedings{suzgun-etal-2023-challenging,
    title = "Challenging {BIG}-Bench Tasks and Whether Chain-of-Thought Can Solve Them",
    author = {Suzgun, Mirac  and
      Scales, Nathan  and
      Sch{\"a}rli, Nathanael  and
      Gehrmann, Sebastian  and
      Tay, Yi  and
      Chung, Hyung Won  and
      Chowdhery, Aakanksha  and
      Le, Quoc  and
      Chi, Ed  and
      Zhou, Denny  and
      Wei, Jason},
    editor = "Rogers, Anna  and
      Boyd-Graber, Jordan  and
      Okazaki, Naoaki",
    booktitle = "Findings of the Association for Computational Linguistics: ACL 2023",
    month = jul,
    year = "2023",
    address = "Toronto, Canada",
    publisher = "Association for Computational Linguistics",
    url = "https://aclanthology.org/2023.findings-acl.824/",
    doi = "10.18653/v1/2023.findings-acl.824",
    pages = "13003--13051",
}

@inproceedings{
zhou2023leasttomost,
title={Least-to-Most Prompting Enables Complex Reasoning in Large Language Models},
author={Denny Zhou and Nathanael Sch{\"a}rli and Le Hou and Jason Wei and Nathan Scales and Xuezhi Wang and Dale Schuurmans and Claire Cui and Olivier Bousquet and Quoc V Le and Ed H. Chi},
booktitle={The Eleventh International Conference on Learning Representations },
year={2023},
url={https://openreview.net/forum?id=WZH7099tgfM}
}

@article{liu-etal-2024-lost,
    title = "Lost in the Middle: How Language Models Use Long Contexts",
    author = "Liu, Nelson F.  and
      Lin, Kevin  and
      Hewitt, John  and
      Paranjape, Ashwin  and
      Bevilacqua, Michele  and
      Petroni, Fabio  and
      Liang, Percy",
    journal = "Transactions of the Association for Computational Linguistics",
    volume = "12",
    year = "2024",
    address = "Cambridge, MA",
    publisher = "MIT Press",
    url = "https://aclanthology.org/2024.tacl-1.9/",
    doi = "10.1162/tacl_a_00638",
    pages = "157--173",
}

@misc{brown2024largelanguagemonkeysscaling,
      title={Large Language Monkeys: Scaling Inference Compute with Repeated Sampling}, 
      author={Bradley Brown and Jordan Juravsky and Ryan Ehrlich and Ronald Clark and Quoc V. Le and Christopher Ré and Azalia Mirhoseini},
      year={2024},
      eprint={2407.21787},
      archivePrefix={arXiv},
      primaryClass={cs.LG},
      url={https://arxiv.org/abs/2407.21787}, 
}

@misc{wu2025roletemperaturesamplingtesttime,
      title={On the Role of Temperature Sampling in Test-Time Scaling}, 
      author={Yuheng Wu and Azalia Mirhoseini and Thierry Tambe},
      year={2025},
      eprint={2510.02611},
      archivePrefix={arXiv},
      primaryClass={cs.AI},
      url={https://arxiv.org/abs/2510.02611}, 
}

@misc{ju2026reasoningpathdivergencenew,
      title={Reasoning Path Divergence: A New Metric and Curation Strategy to Unlock LLM Diverse Thinking}, 
      author={Feng Ju and Zeyu Qin and Rui Min and Zhitao He and Lingpeng Kong and Yi R. Fung},
      year={2026},
      eprint={2510.26122},
      archivePrefix={arXiv},
      primaryClass={cs.CL},
      url={https://arxiv.org/abs/2510.26122}, 
}

@misc{yue2025doesreinforcementlearningreally,
      title={Does Reinforcement Learning Really Incentivize Reasoning Capacity in LLMs Beyond the Base Model?}, 
      author={Yang Yue and Zhiqi Chen and Rui Lu and Andrew Zhao and Zhaokai Wang and Yang Yue and Shiji Song and Gao Huang},
      year={2025},
      eprint={2504.13837},
      archivePrefix={arXiv},
      primaryClass={cs.AI},
      url={https://arxiv.org/abs/2504.13837}, 
}

@inproceedings{
snell2025scaling,
title={Scaling {LLM} Test-Time Compute Optimally Can be More Effective than Scaling Parameters for Reasoning},
author={Charlie Victor Snell and Jaehoon Lee and Kelvin Xu and Aviral Kumar},
booktitle={The Thirteenth International Conference on Learning Representations},
year={2025},
url={https://openreview.net/forum?id=4FWAwZtd2n}
}

@inproceedings{
madaan2023selfrefine,
title={Self-Refine: Iterative Refinement with Self-Feedback},
author={Aman Madaan and Niket Tandon and Prakhar Gupta and Skyler Hallinan and Luyu Gao and Sarah Wiegreffe and Uri Alon and Nouha Dziri and Shrimai Prabhumoye and Yiming Yang and Shashank Gupta and Bodhisattwa Prasad Majumder and Katherine Hermann and Sean Welleck and Amir Yazdanbakhsh and Peter Clark},
booktitle={Thirty-seventh Conference on Neural Information Processing Systems},
year={2023},
url={https://openreview.net/forum?id=S37hOerQLB}
}

@misc{abdin2025phi4reasoningtechnicalreport,
      title={Phi-4-reasoning Technical Report}, 
      author={Marah Abdin and Sahaj Agarwal and Ahmed Awadallah and Vidhisha Balachandran and Harkirat Behl and Lingjiao Chen and Gustavo de Rosa and Suriya Gunasekar and Mojan Javaheripi and Neel Joshi and Piero Kauffmann and Yash Lara and Caio César Teodoro Mendes and Arindam Mitra and Besmira Nushi and Dimitris Papailiopoulos and Olli Saarikivi and Shital Shah and Vaishnavi Shrivastava and Vibhav Vineet and Yue Wu and Safoora Yousefi and Guoqing Zheng},
      year={2025},
      eprint={2504.21318},
      archivePrefix={arXiv},
      primaryClass={cs.AI},
      url={https://arxiv.org/abs/2504.21318}, 
}

@inproceedings{al2025evaluating,
  title={Evaluating Intermediate Reasoning of Code-Assisted Large Language Models for Mathematics},
  author={Al Khalili, Zena and Howell, Nick and Klakow, Dietrich},
  booktitle={Proceedings of the Fourth Workshop on Generation, Evaluation and Metrics (GEM$^2$)},
  pages={741--758},
  year={2025}
}

@inproceedings{liu2025pricinglogic,
  title={PricingLogic: Evaluating LLMs Reasoning on Complex Tourism Pricing Tasks},
  author={Liu, Yunuo and Zhu, Dawei and Al-Khalili, Zena and Cheng, Dai and Chen, Yanjun and Klakow, Dietrich and Zhang, Wei and Shen, Xiaoyu},
  booktitle={Proceedings of the 2025 Conference on Empirical Methods in Natural Language Processing},
  pages={7725--7734},
  year={2025}
}

@inproceedings{
rein2024gpqa,
title={{GPQA}: A Graduate-Level Google-Proof Q\&A Benchmark},
author={David Rein and Betty Li Hou and Asa Cooper Stickland and Jackson Petty and Richard Yuanzhe Pang and Julien Dirani and Julian Michael and Samuel R. Bowman},
booktitle={First Conference on Language Modeling},
year={2024},
url={https://openreview.net/forum?id=Ti67584b98}
}

@inproceedings{aggarwal-etal-2023-lets,
    title = "Let{'}s Sample Step by Step: Adaptive-Consistency for Efficient Reasoning and Coding with {LLM}s",
    author = "Aggarwal, Pranjal  and
      Madaan, Aman  and
      Yang, Yiming  and
      Mausam",
    editor = "Bouamor, Houda  and
      Pino, Juan  and
      Bali, Kalika",
    booktitle = "Proceedings of the 2023 Conference on Empirical Methods in Natural Language Processing",
    month = dec,
    year = "2023",
    address = "Singapore",
    publisher = "Association for Computational Linguistics",
    url = "https://aclanthology.org/2023.emnlp-main.761/",
    doi = "10.18653/v1/2023.emnlp-main.761",
    pages = "12375--12396",
}

@inproceedings{
lin2025critical,
title={Critical Tokens Matter: Token-Level Contrastive Estimation Enhances {LLM}{\textquoteright}s Reasoning Capability},
author={Zicheng Lin and Tian Liang and Jiahao Xu and Qiuzhi Liu and Xing Wang and Ruilin Luo and Chufan Shi and Siheng Li and Yujiu Yang and Zhaopeng Tu},
booktitle={Forty-second International Conference on Machine Learning},
year={2025},
url={https://openreview.net/forum?id=fnz1g18EdI}
}
\bibliographystyle{colm2026_conference}

\appendix
\section{Evaluation}\label{sec:appendix-evaluation}

\subsection{Preliminary Experiments using MATH500}\label{sec:appendix-preliminary}
The MATH dataset was originally published by \citet{hendrycksmath2021} and consists of 12,5k problems (7,500 training and 5,000 test). 
Later, \citet{lightman2024lets} utilized further 4,5k instaces from the original test set for model training, reserving 500 instances as an uncontaminated test set (MATH500).
\Cref{tab:preliminary} shows our preliminary results using DeepSeek-8B for the MATH500 dataset using Greedy, Parallel Thinking, and \method. 
As can be seen, the performance of the simple, greedy baseline is already very high, achieving an accuracy of $\sim$93\%. 
We conjecture that there is a high chance of dataset contamination---i.e., the model might have seen the test set during training---which seems likely considering the dataset was already released in 2021.
We thus decided to exclude this dataset from our evaluation. 
Regardless, we can see that \method uses 33\% less tokens compared to Parallel Thinking.

\begin{table}[!htb]
  \centering
  \small
  \begin{NiceTabular}{lcccccc}
  \toprule
  \textbf{Model} & \multicolumn{2}{c}{\textbf{Greedy}} & \multicolumn{2}{c}{\textbf{Parallel Thinking}} &  \multicolumn{2}{c}{\textbf{\method}} \\
  \cmidrule(lr){2-3} \cmidrule(lr){4-5} \cmidrule(lr){6-7} 
   & Token  & Acc & Token  & Acc & Token ($\Delta$\%) & Acc \\
  \midrule
  DeepSeek-8B  & \(3.6\)  & \(93\%\) & 119.6    & 93\%    & (-33\%) & 94\% \\
  \bottomrule
  \end{NiceTabular}

  \caption{Preliminary results on the MATH500 dataset.}
  \label{tab:preliminary}
\end{table}

\subsection{Hyper-parameters}\label{sec:appendix-parameters}
For each model, we select the hyper-parameters for generation as per recommendation in the respective model card.
We set following hyper-parameters: the maximum trace length $l_{\text{max}}$, the temperature $\tau$, and $p$ for top-$p$ (nucleus) sampling.   
\Cref{tab:04-parameters} provides a list of all model-specific hyper-parameters.

\begin{table}[!htb]
  \centering
  \small
  \begin{NiceTabular}{lccc}
  \toprule
  Model & $l_{\text{max}}$ & $\tau$ & $p$ \\
  \midrule
  Qwen3-8B & 32,000 & 0.7 & 0.95  \\
  DeepSeek-8B & 64,000 & 0.7 & 0.95 \\
  Phi-4-RP-14B & 30,000 & 0.8 & 0.95 \\
  \bottomrule
  \end{NiceTabular}

  \caption{Hyper-parameters used for each model.}
  \label{tab:04-parameters}
\end{table}

\subsection{Run-time Analysis}\label{sec:appendix-runtime}

We further compare the overall run-time of \method and Parallel Thinking to illustrate the practical behavior of our implementation, and find that \method leads to a substantial decrease in terms of overall run-time.
As \cref{tab:04-runtime} shows, the average run-times vary substantially between different models. 
We further find that the run-time reduction is substantially larger when using two GPUs although \method reduces the token cost far less, namely, 14\% compared to the 27\% for Qwen3-8B. 
This is a very promising finding but remains to be investigated in future work using more models and GPUs.

\begin{table}[!htb]
  \centering
  \small
  \begin{NiceTabular}{lrccc}
  \toprule
  Model & \# GPUs & Parallel Thinking & \method & Time ($\Delta$\%)  \\
  \midrule
  Qwen3-8B & 1 & 11.98 & 7.39 & -38\% \\
  DeepSeek-8B & 2 & 20.81 & 8.92 & -57\% \\
  Phi-4-RP-14B & 2  & 13.45 & 6.45 & -52\% \\
  \bottomrule
  \end{NiceTabular}
  \caption{Average run-time of different methods (in hours) on the AIME25 dataset on NVIDIA H100 GPUs. We can see that \method saves up to 57\% of Parallel Thinking run-time using the same compute infrastructure. Note, that the different backends (vLLM in DeepConf and SGLang in \method) renders both methods incomparable with respect to run-time. For completeness, we report the run-time for DeepConf in \cref{tab:04-runtime-deepconf}.}
  \label{tab:04-runtime}
\end{table}

For completeness, \cref{tab:04-runtime-deepconf} provides further run-times for DeepConf but note that these are inherently not comparable due to the different backends. 
We can see that in general, DeepSeek-8B requires the longest amount of time despite having less parameters than Phi-4-RP-14B. 
The differences between Qwen3-8B and DeepSeek-8B can be attributed to the different number of GPUs used for the experiments. 

\begin{table}[!htb]
  \centering
  \small
  \begin{NiceTabular}{lcr}
  \toprule
  Model & \# GPUs & Time   \\
  \midrule
  Qwen3-8B & 1 & 13.42  \\
  DeepSeek-8B & 1 & 27.43 \\
  Phi-4-RP-14B & 1 & 41.6 \\
  \bottomrule
  \end{NiceTabular}

  \caption{Average run-time of DeepConf (in hours) on the AIME25 dataset on NVIDIA A100 (40GB) GPUs.}
  \label{tab:04-runtime-deepconf}
\end{table}

\subsection{Compute Infrastructure} 
The experiments were conducted on a high-performance computing cluster equipped with NVIDIA H100 (80GB) or A100 (40GB/80GB) GPUs. 
GPU usage was limited to at most 8 GPUs being occupied simultaneously, with a single experiment using up to 2 GPUs at once.


\section{Prompt Templates}\label{sec:appendix-prompts}
We use the same prompts across all models and methods. 
To generate our greedy seed path (as well as for the greedy baseline), we use a basic CoT prompt:

\begin{tcolorbox}[
    colframe=paired-light-gray!40!black,
    colback=paired-light-gray!5,
    coltitle=white,
    fonttitle=\bfseries,
    title=Seed path,
    colbacktitle=paired-light-gray!40!black
]
Please reason about the problem step by step. \{Question\} 

\end{tcolorbox}

For Parallel Thinking, DeepConf, and \method, we use the prompt that is stated in the usage recommendations for mathematical problems of the DeepSeek--R1 series:\footnote{\url{https://huggingface.co/deepseek-ai/DeepSeek-R1/blob/main/README.md\#usage-recommendations}}

\begin{tcolorbox}[
    colframe=paired-light-gray!40!black,
    colback=paired-light-gray!5,
    coltitle=white,
    fonttitle=\bfseries,
    title={\method, DeepConf, Parallel Thinking, CoT, ASC},
    colbacktitle=paired-light-gray!40!black
]
Please reason step by step, and put your final answer within boxed\{\}. \{Question\} 

\end{tcolorbox}

\section{Additional Results on Forking Metrics}\label{sec:appendix-FM}

\Cref{tab:parallel-min-max-entropy-aime} provides the results of all models across both AIME datasets for different model confidence estimates (cf.~\S\ref{sec:06-results} for a description). 
We can see that choosing forking points corresponding to lowest average log-probability ($\arg\min$) can substantially save more tokens.
At the same time, the mixed accuracy trends indicate that identifying optimal forking points remains a challenging problem and requires further investigation in future work.

\begin{table}[t]
\centering
\small
\resizebox{\textwidth}{!}{
\begin{tabular}{llcccccccc}
\toprule
Dataset & Model 
& \multicolumn{2}{c}{Parallel Think} 
& \multicolumn{2}{c}{$\arg\min_{10}$} 
& \multicolumn{2}{c}{$\arg\max_{10}$}
& \multicolumn{2}{c}{Entropy$_{1000}$} \\
\cmidrule(lr){3-4} \cmidrule(lr){5-6} \cmidrule(lr){7-8} \cmidrule(lr){9-10}
& & Tok. & Acc.
  & Tok. & Acc. 
  & Tok. & Acc.
  & Tok. & Acc. \\
\midrule
{AIME24}
& Qwen3  & 20.21 & 80.67 & \textbf{14.23} & 81.80 & 17.89 & 81.11 & 18.77 & \textbf{84.44} \\
& DeepSeek & 22.62 & 86.0 & \textbf{20.14} & 86.0 & 21.13 & 86.0 & 21.11 & 86.0 \\
& Phi      & 12.22 & \textbf{77.80} & \textbf{11.38} & 77.33 & 11.54 & 67.78 & 11.61 & 74.44 \\
\midrule
{AIME25}
& Qwen3    & 22.08 & 75.40 & \textbf{15.96} & 75.60 & 20.77 & \textbf{75.66} & 21.47 & 74.44 \\
& DeepSeek & 25.17 & \textbf{83.33} & \textbf{21.92} & 82.41 & 22.99 & 80.00 & 23.22 & 80.0 \\
& Phi      & 13.59 & 62.60 & 12.87 & \textbf{66.67} & \textbf{12.59} & 65.56 & 12.91 & 65.56 \\
\bottomrule
\end{tabular}
}
\caption{Average total tokens (in $\times 10^{6}$) and majority-voting accuracy (in \%) across repeated runs, using a branching factor of $n=32$. $\argmin$ and $\argmax$ are over the average log-probability of the top-$k$ tokens at each position. }
\label{tab:parallel-min-max-entropy-aime}
\end{table}

\end{document}